\documentclass[journal]{IEEEtran}

\usepackage{amsmath,amssymb,amsfonts}
\newtheorem{definition}{Definition}
\usepackage{hyphenat}
\usepackage{textcomp}
\usepackage{graphicx}
\usepackage{subfigure}
\usepackage{url}
\usepackage{hyperref}
\usepackage{array}      
\usepackage{etoolbox}   

\newcounter{mylink}

\newcommand{\numberedlink}[2]{%
    \stepcounter{mylink}%
    \phantomsection\label{link\themylink}
    \href{#2}{\textsuperscript{\themylink}#1}%
}

\makeatletter
\patchcmd{\textsuperscript}{\m@th$^$}{\m@th$^$\kern-0.05em}{}{}
\makeatother

\makeatletter
\newcommand{\customsize}{\fontsize{\dimexpr0.7\dimexpr\f@size pt}{0.7\dimexpr\f@size pt}\selectfont}
\makeatother
\usepackage{tabularx}
\usepackage[numbers,sort&compress]{natbib}
\usepackage{filecontents,url}

\begin{document}
\title{Anomaly Detection in Graph Structured Data: A Survey} 

\author{\IEEEauthorblockN{Prabin~B~Lamichhane\IEEEauthorrefmark{1}, William Eberle\IEEEauthorrefmark{2}}
\IEEEauthorblockA{\\Department of Computer Science\\
Tennessee Tech University\\ Cookeville, TN 38501 USA\\ plamichh@gmail.com\IEEEauthorrefmark{1}, weberle@tntech.edu\IEEEauthorrefmark{2}}
}

\maketitle

\begin{abstract}
Real-world graphs are complex to process for performing effective analysis, such as anomaly detection. However, recently, there have been several research efforts addressing the issues surrounding graph-based anomaly detection. In this paper, we discuss a comprehensive overview of anomaly detection techniques on graph data. We also discuss the various application domains which use those anomaly detection techniques. We present a new taxonomy that categorizes the different state-of-the-art anomaly detection methods based on assumptions and techniques. Within each category, we discuss the fundamental research ideas that have been done to improve anomaly detection. We further discuss the advantages and disadvantages of current anomaly detection techniques. Finally, we present potential future research directions in anomaly detection on graph-structured data.
\end{abstract}


\section{Introduction}\label{intro_ch2}
Many real-world data sets are large, providing a vast array of information. As a result, recently, there has been a growing interest in the ability to analyze information presented in graph form to discover interesting patterns and anomalies. In short, abnormal instances are known as anomalies, and detecting such instances is called \textit{anomaly detection}. Anomaly detection (sometimes called outlier detection) involves discovering rare occurrences that arouse suspicions by deviating significantly from the majority of other occurrences \cite{Zimek}. The anomaly detection problem has numerous high-impact applications in real-world graph-structured datasets. To detect anomalies in networks, we need to account for inter-dependencies (relations and connectivity patterns) in the graph representation \cite{NovelGB}. However, it is challenging to preserve the inter-dependencies of the graph \textit{and} apply a practical analysis. Recent research has presented various well-studied anomaly detection techniques on graph-structured application domains.

\begin{figure}[htbp] 
    \includegraphics[width=8.5cm, keepaspectratio=true]{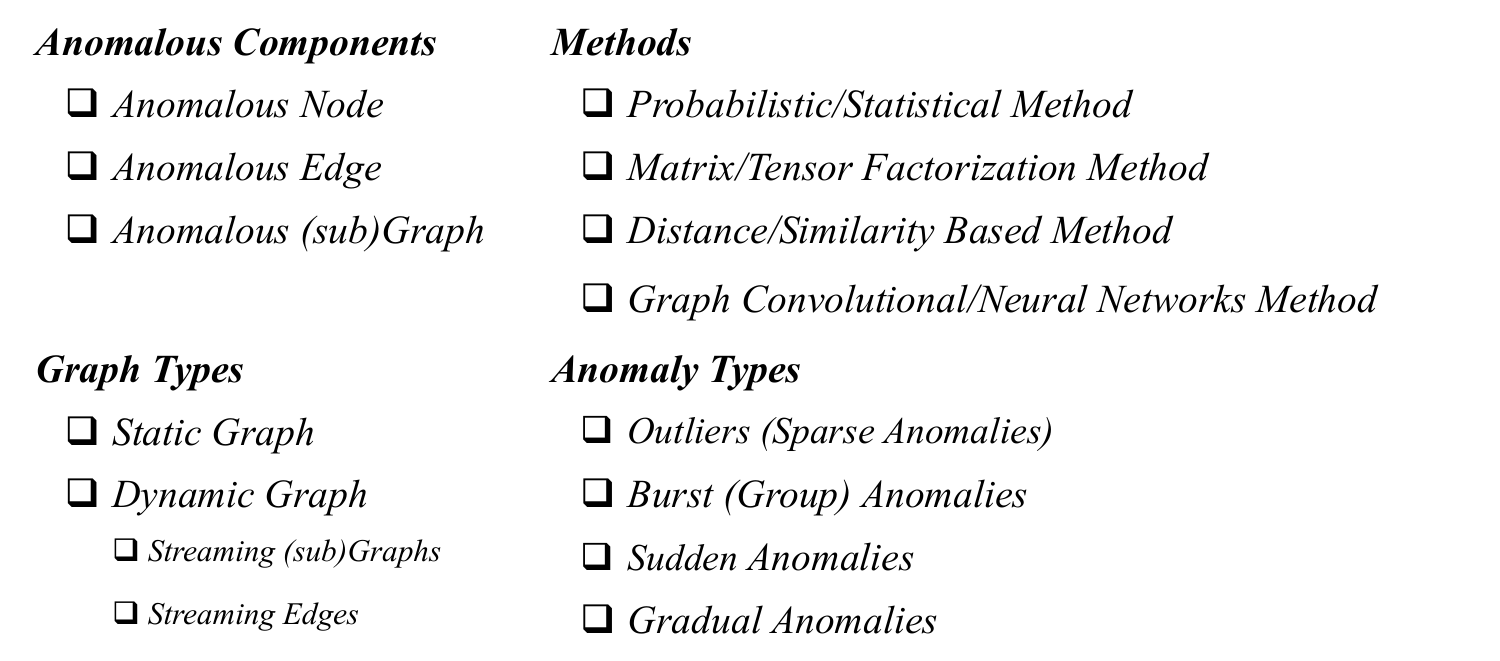}
\caption{Categorization of Anomaly Detection Problem Based on Various Contexts}
\label{Cat}
\end{figure}

The following is a literature survey on existing anomaly detection problems in graph data based on various contexts: type of anomalous component, type of graph, type of method, and application areas (types of anomalies), as shown in Figure \ref{Cat}. 

This work is organized as follows: section \ref{rw_ch2} presents an overview of surveys on anomaly detection. Section~\ref{tac} categorizes the anomaly detection problem in graph-structured data based on anomalous entities. In sections \ref{tgd}, \ref{tgt}, \ref{ta}, and \ref{app} we categorize anomaly detection papers based on types of graph data, types of graph-based techniques, types of anomaly, and applications, respectively. Then, we present challenges and future directions of graph-based anomaly detection methods in section \ref{cfd}. Finally, we conclude with some remarks in section \ref{con}.
\section{Overview of Surveys on Anomaly Detection} \label{rw_ch2}
Several literature surveys have been performed on anomaly detection \cite{AgAg2015, AhNa2016, AhNa2019, MoHu2019} and anomaly detection using graph-based techniques \cite{SaZh2014, RaSh2015, AkTo2015, RoHu2016, KaSi2016, CaCh2018, PoOn2020}. However, none of these surveys have provided a detailed exploration of graph-based techniques in anomaly detection. In this work, we provide a comprehensive overview of various state-of-art anomaly detection techniques in graph-structured data. We categorize anomaly detection papers based on the type of anomalous component, graph type, method used, and anomaly type. We categorize research papers undertaken in the last two decades to provide a consolidated understanding of anomaly detection problems on graphs. We further discuss the challenges raised in each context and solutions to resolve them. Additionally, we present future directions to expand anomaly detection research on graph-structured data.
\section{Types of Anomalous Component}\label{tac}
There are three basic types of anomalies in graph networks. These anomaly types represent the various components of a graph. For instance, in computer networks, the computer devices are represented by \textit{nodes}, the connections/communications between devices are represented by \textit{edges}, and the sub-networks/communities of computer devices are represented as \textit{(sub)graphs}. For example, a specific attack - an illicit connection established between an attack device and a normal device - could be a sub-network of devices more abnormal than a regular sub-network, represented as an anomalous node, edge, and sub-graph, respectively. We define a graph $\mathcal{G} = (\mathcal{V}, \mathcal{E})$, where $\mathcal{V}$ is the set of nodes and $\mathcal{E} \subseteq (\mathcal{V} \times \mathcal{V})$ is the set of edges between nodes. An edge $e_{i,j,w} \in \mathcal{E}$ exists if the node $i$ is connected with node $j$ with its weight $w$, representing the number of connections. We describe each anomalous component in more detail in the following subsections.

\subsection{Anomalous Nodes} 
\begin{figure}[htbp]
    \subfigure []{\includegraphics[width=4cm]{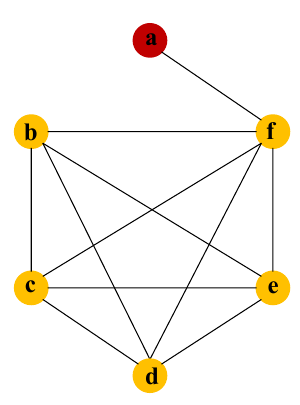}}
    \hspace{5mm}
    \subfigure []{\includegraphics[width=4cm] {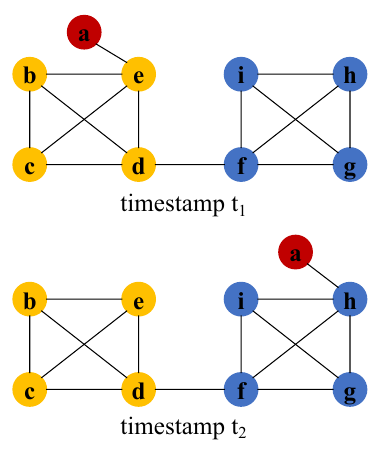}}
\caption{Anomalous node in (a) static, and (b) dynamic graph.}
\label{AN}
\end{figure}

Nodes in a graph are said to be anomalous if they are unexpected compared to other nodes in the graph. The unusual nodes could consist of an unusual degree of connections, unexpected community nodes, unexpected labels, etc., depending upon whether the graph is static or dynamic. For example, in static graphs, one might compare nodes with each other based on the degree of connection with neighboring nodes within the graph snapshot. In Figure~\ref{AN}(a), one can see that all nodes (except node $a$, denoted as a red node) on the snapshot have a degree of 4 or more, whereas node $a$ has a degree of 1. This makes node $a$ anomalous compared to other nodes. However, in dynamic graphs, one would consider the temporal aspect to detect a type of anomaly that is not present in the static graph. In Figure~\ref{AN}(b), one can see two graph snapshots at different timestamps $t_1$ and $t_2$, where node $a$ (red node) at timestamp $t_2$ is anomalous compared to the timestamp $t_1$ as it changes community (i.e., from yellow community nodes to blue community nodes).

\subsection{Anomalous Edges} 
\begin{figure}
    \subfigure[]{\includegraphics[width=4cm]{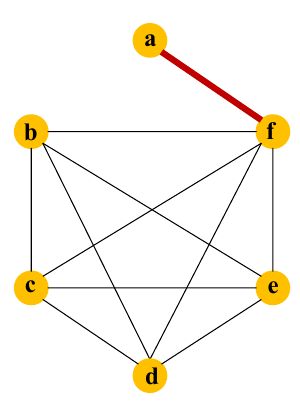}}
    \hspace{3mm}
    \subfigure[]{\includegraphics[width=4cm]{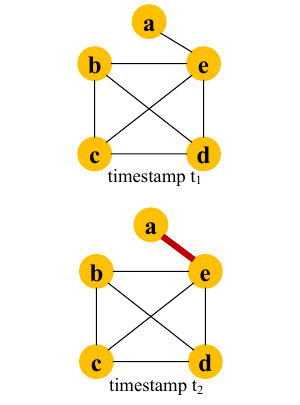}}  
\caption{An anomalous edge in (a) static, and (b) dynamic graph}
\label{AE}
\end{figure}

Similar to anomalous nodes, edges are said to be anomalous if the edges show unusual patterns compared to the other edges in the graph. In a static graph, an edge appears to be anomalous if the edge's weight is abnormal compared to the distribution of the edge weights in the graph. In Figure~\ref{AE}(a), the edge between nodes $a$ and $f$ (denoted by the thick-dark red line - the higher the weight the thicker the line in these visualizations) has a high weight (i.e., high frequent (burst) connections) compared to the weight of other edges in the graph. However, in dynamic graphs, as the temporal aspect is also considered, two types of anomalous edge problems can be detected: 1. like in a static graph, the edge with abnormal weight compared to the weight of other edges in a particular graph snapshot (i.e., timestamp) is considered anomalous, and 2. the edge with abnormal weight in a particular snapshot (i.e., timestamp) compared to the weight of the same edges in other graph snapshots (i.e., timestamps) is also considered anomalous. In Figure~\ref{AE}(b), the edge $e_{ae}$ between nodes $a$ and $e$ has a normal weight like the weights of other edges at timestamp $t_1$, however, at timestamp $t_2$, the edge $e_{ae}$ (denoted by the thick-dark red line) has a higher than expected weight. Thus, when we compare the weights of the edge $e_{ae}$ at the timestamps $t_1$ and $t_2$, the edge $e_{ae}$ at $t_2$ is considered anomalous. 

\subsection{Anomalous (sub)Graphs} 
\begin{figure}[htbp]
    \subfigure[]{
    \includegraphics[width=4cm]{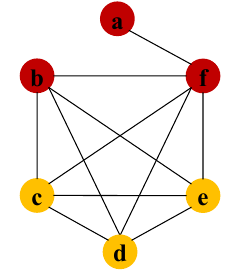}}
    \hspace{3mm}
    \subfigure[]{
    \includegraphics[width=4cm]{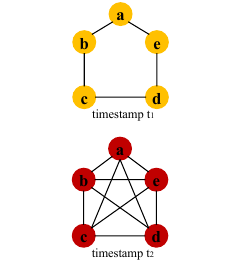}}  
\caption{An anomalous (sub)graph in (a) static, and (b) dynamic graph.}
\label{AG}
\end{figure}

A sub(graph) is considered anomalous if that substructure or sub-graph has an unexpected pattern when compared to other (sub)graphs. Most community detection methods keep track of sub-graph structures and compare similarities to detect anomalous sub-graphs. The graph is divided into different sub-graphs for static graphs, and their structural representations are compared to determine if two sub-graphs are anomalous or not. In Figure~\ref{AG}(a), the initial graph is divided into sub-graphs, each having three nodes, such as  $bcd$, $bde$, $cde$, $dfe$, and $afb$.  Using a vector of the degree of each node in the sub-graph to represent the sub-graph, $bcd$, $bde$, $cde$, $dfe$, and $afb$ sub-graphs can be represented by $[4, 4, 4]$, $[4, 4, 4]$, $[4, 4, 4]$, $[4, 4, 4]$  and $[1, 4, 4]$ respectively. One can see that the $afb$ sub-graph (denoted by red nodes) has an abnormally different representation than other sub-graphs. However, in a dynamic graph, similar to the anomalous nodes/edges presented in the previous sections, one can also keep track of the evolution of sub-graphs at different timestamps to detect temporal anomalous sub-graphs. In Figure~\ref{AG}(b), notice that all nodes have regular connections at timestamp $t_1$. However, at timestamp $t_2$, there will be the addition of several links, drastically changing the graph's structure. The graph structure at timestamp $t_2$ (denoted by red nodes) appears abnormal compared to the graph at timestamp $t_1$. 

\section{Types of Graph Data}\label{tgd}
A graph is a powerful technique to represent a network's inter-dependencies (relations/connections). In the last two decades, there has been growing interest in detecting anomalies in graphs because of the ability to represent inter-dependencies.

\subsection{Anomaly Detection in Static Graphs} \label{ADSG}
Originally, anomaly detection problems were focused on static graphs (that do not evolve with time). This section will discuss various anomaly detection research in static graphs. We formally define a static graph as follows: 
\hfill \break
\begin{definition}
 A static graph (i.e., a graph snapshot) $\mathcal{G}$ can be represented by $\mathcal{G} = (\mathcal{V}, \mathcal{E})$, where $\mathcal{V}$ is the set of vertices and $\mathcal{E} \subseteq (\mathcal{V}\times \mathcal{V})$ is the set of edges between vertices. An edge $e_{i,j} \in \mathcal{E}$ exists if the node $i$ is connected with node $j$.
\end{definition}
\hfill \break
\indent The primary goal of anomaly detection techniques on static graphs is to detect anomalous nodes, edges, and/or sub-graphs that deviate significantly from the observed patterns. \cite{AkMc2010, GaLi2010, SuHu2010, HeGa2011, DaLi2011, ToLi2012, DiKa2012, MuSa2013, GuMa2014, PeAk2014, JiCu2016, PeAk2016, HoSh2017, ShEl2018} are some static approaches to detecting anomalies in graph data. They are primarily what are described as \textit{offline} anomaly detection techniques. That is, this approach considers the network to be a static graph and detects anomalous entities within that graph. However, many real-world networks, such as computer networks, social networks, IoT networks, etc., are dynamic (i.e., the graph changes with time). As time progresses, the size of the network (graph) changes as new nodes/edges are added or nodes/edges are deleted. In addition, static approaches do not consider the graph's dynamic (temporal) aspects while detecting anomalies. Thus, static approaches are not suitable for time-evolving graphs.


\subsection{Anomaly Detection in Dynamic Graphs}
So, while static approaches are applied on graphs that do not change with time and are limited in size, real-world networks evolve with time and are more commonly represented in a streaming fashion \cite{aggarwal2011classification,chi2013fast}. In this section, we present an overview of various anomaly detection research in dynamic graphs by dividing them into two main categories.

\subsubsection{Streaming Graphs}
For streaming graph techniques, the streaming of edges is aggregated into graph snapshots, which are later processed to detect anomalies. We formally define streaming graphs as follows:  
\hfill \break
\begin{definition}
A graph stream $\mathcal{G}$ is the set of infinite sequence of graphs, $\mathcal{G} =\{G_t\}_{t=1}^{\infty}$, where $G_t = (V_t,E_t)$ represents a static graph that contains the set of nodes $V_t$ and the set of edges $E_t \subseteq (V_t\times V_t)$ at timestamp $t$. An edge $e_{i,j,w} \in E_t$ exists if the node $i$ is connected with node $j$ with its weight $w$ at timestamp $t$. 
\end{definition}
\hfill \break
\indent \cite{SuTa2006, SuFa2007, GuGa2012, MaMa2013, BeXu2013, SrDa2014, WaTa2014, PeCl2015, MaMi2016, KoSh2016, TeLi2017, YuAgWa2017, EsFaGu2018, ZhLi2019, YoHo2019, PaEb2020} are some streaming graph approaches for detecting anomalies in dynamic graphs, and are generally referred to as \textit{online} anomaly detection techniques. These approaches accumulate the edges for a certain time period and form a graph snapshot. Later, they process the graph snapshots formed at different time periods to detect anomalous entities. The graph streaming methods consume significantly more time to detect anomalies as they need to collect streaming edges to form graph snapshots. However, there is still a need to detect anomalies as soon as possible to reduce the effect of anomalous activities. 

\subsubsection{Streaming Edges}
In contrast, a streaming edges anomaly detection technique processes streams of edges over time. We formally define streaming edges as follows:
\hfill \break
\begin{definition}
A edge stream $\mathcal{E}$ is the set of infinite sequence of edges, $\mathcal{E} =\{e_i\}_{i=1}^{\infty} = \{e_1, e_2, e_3, \dots \}$ from a dynamic graph $\mathcal{G}$, where each edge $e_{i} = (s_i, d_i, w_i, t_i)$ exists if source node $s_i$ is connected with destination node $d_i$ with its weight $w_i$ at time tick $t_i$. 
\end{definition}
\hfill \break
\indent \cite{YuAg2013, ShHo2017, MaMo2016, RaHa2016, EsFa2018, YuCh2018, BhHo2020, BeZh2020, ChLi2021} are some streaming edges approaches to detect anomalies in dynamic graphs. These approaches are also online anomaly detection techniques. Unlike streaming graphs, this technique doesn't accumulate edges but instead processes the edge as it appears and detects anomalies near real-time. That is, edge streaming techniques are used to detect anomalies as soon as possible so that the remedy can be applied before a large disaster occurs. 

However, anomaly detection in dynamic graphs is challenging for the following reasons: 
\begin{itemize}
\item \textbf{Increasing graph volumes:} The edges arrive continuously in the stream, leading to growing graph data. 
\item \textbf{Expanding feature space:} The graphs in the stream are received in real-time, leading to an increasing number of features (like subgraph, clique, subtree, etc.) patterns. 
\item \textbf{One pass requirement:} The graph in the stream is available only once. 
\end{itemize}

All of these challenges have to deal with time and space complexities along with anomaly detection accuracy. In recent years, much research has been conducted on these challenges. 
\section{Types of Graph-based Techniques}\label{tgt}
In this section, we briefly describe the different technical foundations used in detecting anomalies in graph-structured data.

\subsection{Probabilistic/Statistical Techniques}
Anomaly detection models that use probabilistic/statistical methods are based on characterizing the normal patterns of a network and checking for deviations from those typical patterns. NetProbe \cite{PaCh2007} uses a Markov Random Field (MRF) to tune data into a graphical form suitable to detect suspicious patterns through statistical analysis of neighboring nodes and a Belief Propagation algorithm to detect fraudsters. Anomaly detection in OddBall \cite{AkMc2010} is related to Egonet patterns (i.e., power laws) discovery: the few nodes that deviate from a pattern are considered outliers. The Egonet is, for each node, the sub-graph of surrounding neighboring nodes. NetSpot \cite{MoBo2013} also considers the temporal changes to detect anomalies in the graphs. It is specifically designed to detect Significant Anomalous Regions (SAR) in snapshots of graphs. First, it initially characterizes the average behavior of edges. Then, it computes the anomaly score of edges at a given time based on the deviation of their behavior from the average. Finally, it uses a neighborhood search approach to extend the regions of anomalous edges. HotSpot \cite{YuAg2013} is designed for graph streams to detect anomalous changes by keeping track of the statistics of a subset of nodes for online analysis (or all nodes, in the case of offline analysis). HotSpot maintains the mean and standard deviations continuously over time to compute the unusual deviations. NDKD \cite{HeWe2010} uses a two-stage method to detect anomalies in dynamic graphs. First, the Bayesian probability model is used to model the normal behavior in the graph by keeping track of the communication link counts. Second, standard network tools like spectral clustering are used to reduce a subset of potential anomalous nodes by characterizing the graph's structure.

\cite{PrCo2005, NeHa2013, ChNe2014, WaTa2014, PeCl2015} use scan statistics to detect unusual or anomalous regions. A scan statistic is a high (or low) statistic of a local region of the data. It is sometimes called "moving window analysis." In each window period, a local statistic is computed and checked for outliers that have substantially higher (or lower) statistics than those seen in the recent past.
\cite{PrCo2005} considers a star shape connectivity in the graph to be a local region. Whereas JCAMC \cite{NeHa2013} considers star (to detect scanning behavior) and k-path (to detect traversal behavior) shapes. All stars or 3-paths are enumerated for each time interval, and their p-values are computed using a Hidden Markov Model (HMM) to detect anomalous regions (if the p-value is less than the threshold). NPHGS \cite{ChNe2014} first models the baseline distribution of a graph based on the neighborhood environment of each node and estimates the empirical p-value to represent the anomalousness of each node for each time interval. Later, it uses non-parametric scan statistics to detect the most anomalous clusters in heterogeneous graphs. LocalityStat \cite{WaTa2014} uses locality-based scan statistics for change-point detection in dynamic graphs. Generalized Hierarchical Random Graph (GHRG) \cite{PeCl2015} uses a parametric probability distribution model to represent community structure in a graph. Later, it uses the Bayesian Hypothesis test with user-defined parameters to detect a change-point. 

\cite{BeXu2013, ShBe2014, JiCu2016, HoSh2017, ShEl2018} use dense-subgraph detection techniques to detect fraud or spam in social and e-commerce networks. CopyCatch \cite{BeXu2013} detects densely connected subgraphs in social networks. For instance, it spots ill-gotten Page Likes by detecting near-bipartite cores (i.e., the same type of users Like the same Pages) with certain edge constraints (i.e., Like frequency). However, small-scale, stealthy attacks may not be detected by CopyCatch. So, F-BOX \cite{ShBe2014} uses an adversarial approach to detect stealth attacks. CatchSYN \cite{JiCu2016} detects suspicious followers and botnets based on the synchronicity (i.e., an occurrence at the same rate) and the abnormality (i.e., unusual pattern with respect to the majority of nodes). It is a parameter-free technique that does not require the side information as CopyCatch. Unfortunately, the smart fraudsters try to appear normal by linking themselves to popular users (like popular actors/singers), and CatchSYN and F-BOX cannot detect this camouflage behavior. However, FRAUDAR \cite{HoSh2017} can detect fraudsters that are camouflaged, and provides theoretical guarantees on detection. All the former methods detect the densest subgraphs, but CoreScope \cite{ShEl2018} detects the smaller dense subgraphs as complementary to them. CoreScope discovers the mirror pattern, where "the vertices with high coreness tend to have high degrees and vice versa." However, anomalies like Loner-Star (high degree, but low coreness), Lockstep Behavior (high coreness, low degree), etc., deviate from the mirror pattern. 

Detecting suspicious lockstep behavior like fake review boosting attacks using dense subgraph detection techniques has been extensively studied. Additional information, such as reviews that occurred simultaneously, having the same stars and keywords, etc., on a dense subgraph makes the reviews more likely to be suspicious. The best way to include such additional information is to transform data into a tensor and then discover dense sub-tensors. In computer networks, one can consider an edge between a source IP and a destination IP as a 2-way tensor (source IP, destination IP). Likewise, pne can consider a stream of edges as a 3-way tensor (source IP, destination IP, timestamp). In e-commerce networks, Ratings on products can be considered a 4-way tensor (user, product, rating, timestamp). In such applications, \cite{JiBe2015, ShHo2016, ShHoKi2017, ShHo2017, BhJa2021} consider sudden dense sub-tensors in tensors to detect network attacks, fake rating attacks, etc. CrossSpot \cite{JiBe2015} detects a suspicious subtensor (i.e., finds and measures the suspiciousness of blocks) in multimodel data. However, CrossSpot is not satisfactory in terms of accuracy, and speed. Also, it lacks density metrics and guarantees of accuracy. So, M-ZOOM \cite{ShHo2016} proposed a flexible framework that considers density metrics to detect dense subtensors in tensors. Both methods, CrossSpot and M-ZOOM, consider the tensors to be small enough to fit into memory which limits their application on large real-world networks like the web or social networks. D-Cube \cite{ShHoKi2017} applies a disk-based dense tensor technique to fit large dense tensors into memory. However, all of these dense subtensor detection methods only consider a static tensor. DenseAlert \cite{ShHo2017} proposed an incremental algorithm that assumes changes in a tensor (i.e., streaming tensor over time) to spot sudden dense subtensors. Most dense subtensor methods consider categorical features as tensors, however, the recent work of MStream \cite{BhJa2021} detects the dense anomalies in a stream of multi-aspect data by considering both the categorical features (such as IP addresses) and the numerical features (such as average packet size).  

\begin{table*}
\caption{Probabilistic/Statistical Techniques}
\begin{tabular}{|p{1.5cm}|p{2cm}|p{1cm}|>{\baselineskip=7pt}p{4.1cm}|>{\baselineskip=7pt}p{3.3 cm}|>{\baselineskip=7pt}p{3.5cm}|}
\hline
\textbf{Anomalous Component} & \textbf{Papers} & 
\textbf{Graph Types} & \textbf{Techniques} & \textbf{Anomalies}& \textbf{Datasets}\\
\hline
\textit{Node} 
& OddBall \cite{AkMc2010} & \textit{Static} & Egonet Patterns (count of triangles, total weight, and principal eigenvalues) & Abnormal nodes in weighted graphs & Postnet, DBLP\hyperref[link3]{\textsuperscript{3}}, Com2Cand\hyperref[link27]{\textsuperscript{27}}, Don2Com\hyperref[link27]{\textsuperscript{27}}, ENRON\hyperref[link1]{\textsuperscript{1}}, Oregon\hyperref[link26]{\textsuperscript{26}}\\ 
\cline{2-6}
& NetProbe \cite{PaCh2007} &  \textit{Dyn.} & Markov Random Fields and belief propagation & E-commerce networks & Synthetic graph, eBay data\\ 
\hline

\textit{Edge} 
& RHSS \cite{RaHa2016} & \textit{Dyn.} & Count min sketching & Deviation from the standard patterns and showing interest in new trends & IMDB\hyperref[link12]{\textsuperscript{12}}, DBLP\hyperref[link3]{\textsuperscript{3}}, ENRON\hyperref[link1]{\textsuperscript{1}}\\
\cline{2-6}
& MIDAS \cite{BhHo2020} & \textit{Dyn.} & Count-min sketching and Statistical (chi-squared test) & Lockstep behaviors, DoS attacks & DARPA\hyperref[link4]{\textsuperscript{4}}, TwitterSecurity\hyperref[link14]{\textsuperscript{14}}, TwitterWorldCup\hyperref[link17]{\textsuperscript{17}}\\
\cline{2-6}
& PENminer \cite{BeZh2020} & \textit{Dyn.} & Persistent activity snippets (i.e., reoccurring sequences of edge-updates) mining & Both subtle and bursty activities in social or road networks & Eu Email network\hyperref[link5]{\textsuperscript{5}}, Bike network, NYCTaxi\hyperref[link6]{\textsuperscript{6}}, DARPA\hyperref[link4]{\textsuperscript{4}}, Reddit\hyperref[link28]{\textsuperscript{28}}, Stackoverflow\hyperref[link29]{\textsuperscript{29}}\\
\cline{2-6}
& TF-IGF \cite{LaEb2021} & \textit{Dyn.} & Count-min sketching and a variant of TF\-IDF & Lockstep behaviors, DoS attacks & CTU-13\hyperref[link8]{\textsuperscript{8}}, UNSW-NB15\hyperref[link7]{\textsuperscript{7}}, DARPA\hyperref[link4]{\textsuperscript{4}}, ENRON\hyperref[link1]{\textsuperscript{1}}, TwitterSecurity\hyperref[link14]{\textsuperscript{14}}\\
\cline{2-6}
& Isconna \cite{LiSi2021} & \textit{Dyn.} & Count min sketching + G-test statistic & Detecting sudden bursts and pattern deviations & CIC-DDoS\hyperref[link25]{\textsuperscript{25}}, CIC-IDS\hyperref[link25]{\textsuperscript{25}}, DARPA\hyperref[link4]{\textsuperscript{4}}, CTU-13\hyperref[link8]{\textsuperscript{8}}, UNSW-NB15\hyperref[link7]{\textsuperscript{7}}, ISCX-IDS\hyperref[link25]{\textsuperscript{25}}\\
\cline{2-6}
& MIDAS-F \cite{BhLi2022} & \textit{Dyn.} & Count-min sketching and Statistical (chi-squared test) & Lockstep behaviors, DoS attacks & DARPA\hyperref[link4]{\textsuperscript{4}}, CTU-13\hyperref[link8]{\textsuperscript{8}}, UNSW-NB15\hyperref[link7]{\textsuperscript{7}}, TwitterSecurity\hyperref[link14]{\textsuperscript{14}}, TwitterWorldCup\hyperref[link17]{\textsuperscript{17}} \\
\hline

\textit{(Sub)Graph} 
& FBOX \cite{ShBe2014} & \textit{Static} & Adversarial method & Lockstep behaviors (stealth attacks and camouflage resistant) & Twitter\hyperref[link15]{\textsuperscript{15}}, Amazon\hyperref[link30]{\textsuperscript{30}}\\
\cline{2-6}
& CatchSYNC \cite{JiCu2016} & \textit{Static} & Measure synchronicity and normality (degree values and HITS score) & Lockstep behaviors, DoS attacks & TWITTER\hyperref[link15]{\textsuperscript{15}}, WEIBO\hyperref[link16]{\textsuperscript{16}}\\
\cline{2-6}
& FRAUDAR \cite{HoSh2017} & \textit{Static} & Suspiciousness metrics (the node suspiciousness and the edge suspiciousness)  & Fraudsters (camouflage or hijacked accounts) & Amazon\hyperref[link71]{\textsuperscript{71}}, Trip Advisor, EPINIONS\hyperref[link18]{\textsuperscript{18}}, Wiki-vote\hyperref[link31]{\textsuperscript{31}}\\
\cline{2-6}
& CoreScope \cite{ShEl2018} & \textit{Static} & MIRROR PATTERN (the coreness of a vertex, and correlated to vertex degree) & Loner-stars, Lockstep behaviors & (Friendster, Orkut, Flickr, YouTube, Catster\, Skitter, Email, Twitter, LiveJournal, Patent)\hyperref[link32]{\textsuperscript{32}}\\
\cline{2-6}
& ScanStat \cite{PrCo2005} & \textit{Dyn.} & Scan a small window over the data, and calculate local statistics & Detect subdigraphs with an unusually high connectivity & ENRON\hyperref[link1]{\textsuperscript{1}}\\
\cline{2-6}
& NDKD \cite{HeWe2010} & \textit{Dyn.} & Bayesian models and Standard network inference tools & Detects more or less frequently communicating nodes & VAST\hyperref[link33]{\textsuperscript{33}}, EMM network \\
\cline{2-6}
& JCAMC \cite{NeHa2013} & \textit{Dyn.} & Locality statistic, and Maximum scan statistic deviation score over all local area & Burst communications (intruders compromising multiple computers) &  LANL's internal network\\
\cline{2-6}
& NetSpot \cite{MoBo2013} & \textit{Dyn.} & Approximating Heaviest Dynamic Subgraph (HDS) via Maximum Score Subsequence (MSS) and Heaviest Subgraph (HS)& Anomalous event, such as traffic accident, DoS & Small and Large Traffic Networks\hyperref[link41]{\textsuperscript{41}}, Wikipedia, ENRON\hyperref[link1]{\textsuperscript{1}} \\
\cline{2-6}
& HotSpot \cite{YuAg2013} & \textit{Dyn.} & Use fast incremental eigenvector update algorithm & Abrupt or surprising change detection & IMDB\hyperref[link12]{\textsuperscript{12}} \\
\cline{2-6}
& CopyCatch \cite{BeXu2013} &  \textit{Dyn.} & Provably convergent algorithm and MapReduce & Social networks (suspicious lockstep behavior)& Synthetic data, Facebook\hyperref[link32]{\textsuperscript{32}}\\
\cline{2-6}
& LocalityStat \cite{WaTa2014} & \textit{Dyn.} & Stochastic Block Model and Scan statistics constructed from two locality statistics & Change-point detection & ENRON\hyperref[link1]{\textsuperscript{1}} \\
\cline{2-6}
& NPHGS \cite{ChNe2014} & \textit{Dyn.} & Empirical p-value, and Maximum non-parametric scan statistic over connected subgraphs & Event detection and forecasting (such as disease outbreaks, civil unrests, and financial crises) & Twitter data + civil unrest + Rare disease outbreak \\
\cline{2-6}
& GHRG \cite{PeCl2015} & \textit{Dyn.} & Generalized Hierarchical Random Graph (GHRG) and Bayesian hypothesis test with a user-defined parameters & Change-point detection & MIT Reality Mining proximity network\hyperref[link42]{\textsuperscript{42}}, Enron\hyperref[link1]{\textsuperscript{1}} \\
\cline{2-6}
& CrossSpot \cite{JiBe2015} & \textit{Dyn.} & Suspiciousness Score & Lockstep behaviors, DDoS attacks & WEIBO\hyperref[link16]{\textsuperscript{16}}, Synthetic data\\
\cline{2-6}
& M-Zoom \cite{ShHo2016} & \textit{Dyn.} & Suspiciousness Score with Density Measure & Lockstep behaviors, DDoS attacks & (StackOverflow, Youtube, KoWiki, EnWiki, Yelp, Netflix, YahooMessenger, AirForce)\hyperref[link43]{\textsuperscript{43}}\\
\cline{2-6}
& D-Cube \cite{ShHoKi2017} & \textit{Dyn.} & Disk-based dense-block detection & Lockstep behaviors, Bot activities & (Yelp, Android, Netflix, YahooMessenger, Youtube, SMS, KoWiki, EnWiki,  DARPA, AirForce)\hyperref[link44]{\textsuperscript{44}}\\
\cline{2-6}
& MStream \cite{BhJa2021} & \textit{Dyn.} & Tensor based including categorical and numeric attributes & Lockstep behaviors, DDos, DoS & KDDCUP99\hyperref[link13]{\textsuperscript{13}}, CIC-IDS\hyperref[link25]{\textsuperscript{25}}, UNSW-NB15\hyperref[link7]{\textsuperscript{7}} \\
\hline
\end{tabular}
\end{table*}

\subsection{Matrix/Tensor Decomposition Techniques}
Most graph-structured data are multidimensional. A multi-way tensor or matrix is a useful method to represent such data. Matrix analysis, such as Principal Component Analysis (PCA) and Singular Value Decomposition (SVD), or tensor analysis, such as Tensor Decomposition (TD), are popular methods for dimensionality reduction and feature extraction on multi-dimensional data. Matrix/Tensor decomposition techniques have been widely used to detect anomalies in various networks.

Some spectral methods analyze the latent factors produced in matrix decomposition. EigenSpokes \cite{PrSe2009} and GetTheScoop \cite{JiCu2014} are spectral methods to detect suspicious links in social networks. These approaches use the SVD of the graph’s adjacency matrix to detect suspicious behavior. Other anomaly detection techniques use matrix decomposition methods like SVD/PCA, and their variants and extensions. However, they are generally 2-dimensional (e.g., source IP, destination IP), and one often needs to consider other features like timestamps, port number, the number of packets, etc., in computer networks. So, a tensor, a generalization of a matrix, is used to consider multiple aspects of data. Dynamic Tensor Analysis (DTA) \cite{SuTa2006} and its variant, Streaming Tensor Analysis (STA), form a compact synopsis of high-order and high-dimensional data to detect deviations from normal patterns. DTA/STA analyzes the evolution between graph/tensor snapshots that occur in adjacent timestamps. Low-rank decomposition, such as SVD, often ignores the sparsity of graphs. So, the Compact Matrix Decomposition (CMD \cite{SuXi2007}) is used to calculate sparse low-rank approximations. CMD consumes less time and space than SVD. Likewise, DynAnom \cite{MaMa2013} capitalizes on the sparsity and low-rank property of graphs to efficiently process links and track anomalies in near real-time. MET \cite{KoSu2008} uses a variant of Tucker decomposition, Memory-Efficient Tucker (MET), to summarize and analyze graph-structured data. MultiAspectForensics (MAF) \cite{GuMa2011} treats the timestamp as a separate data aspect to detect global anomalies. MAF transforms a heterogeneous network into different tensor modes and performs Canonical Polyadic (CP) decomposition (which generalizes the SVD for matrices) to detect dense substructure patterns. TensorSplat \cite{KoPa2012} uses the "PARAFAC" decomposition method to detect micro-clusters, rare events, changes in patterns, and, in general, anomalies. MTHL \cite{TeLi2017} uses a bilinear dimensionality reduction approach to preserve both temporal and structural features of graphs. Temporal Matrix Factorization (TMF \cite{YuAgWa2017}) uses low-rank properties of the graph that is parameterized by time (i.e., represents graph structure as a function of time). Low-rank factorization methods detect various structural anomalies, however, they can not control p-values to detect distributed attacks (i.e., a group of simultaneous links) properly. Thus, F-FADE \cite{ChLi2021} applies a frequency-factorization technique to detect various anomalies, including distributed attacks. That is, it models the distribution of frequencies of links in a time-evolving graph and computes anomalies based on the maximum likelihood of frequencies of links observed so far. 

\begin{table*}
\caption{Matrix/Tensor Decomposition Techniques}
\begin{tabular}{|p{1.5cm}|p{2.2cm}|p{0.8cm}|>{\baselineskip=7pt}p{4.1cm}|>{\baselineskip=7pt}p{3.3 cm}|>{\baselineskip=7pt}p{3.5cm}|} 
\hline
\textbf{Anomalous Component} & \textbf{Papers} & 
\textbf{Graph Types} & \textbf{Techniques} & \textbf{Anomalies} & \textbf{Datasets} \\
\hline
\textit{Node} 
& DTA/STA \cite{SuTa2006} & \textit{Dyn.} & The dynamic and the streaming tensor Analysis & Port Scanning, DDoS, Abnormal ports & IP2D, IP3D, DBLP\hyperref[link3]{\textsuperscript{3}}\\ 
\hline

\textit{Edge}
& AutoPart \cite{Ch2004} &  \textit{Static} & Lossless MDL-based compression scheme & Outlier Edge Detection & Synthetic (CAVE), DBLP\hyperref[link3]{\textsuperscript{3}}, EPINIONS\hyperref[link18]{\textsuperscript{18}} \\
\cline{2-6}
& NrMF \cite{ToLi2012} &  \textit{Static} & Non-negative residual matrix factorization framework & port-scanner, DDoS, Lockstep behaviors & MIT-DP,
NIPS-PW, CIKM-PA, MovieLens \\
\cline{2-6}
& F-FADE \cite{ChLi2021} & \textit{Dyn.} & Matrix Factorization + Probabilistic measure (max-likelihood) & Communication networks, Financial networks & DARPA\hyperref[link4]{\textsuperscript{4}}, ENRON\hyperref[link1]{\textsuperscript{1}}, DBLP\hyperref[link3]{\textsuperscript{3}}, BARRA\hyperref[link34]{\textsuperscript{34}} \\
\hline

\textit{(Sub)Graph} 
& EigenSpokes \cite{PrSe2009} & \textit{Static} & Spokes (cliques and bipartite cores) detection algorithm & Spotting and extracting tightly-knit communities & Mobile Call graph, Patent\hyperref[link35]{\textsuperscript{35}}, Internet\hyperref[link35]{\textsuperscript{35}}, Dictionary\hyperref[link35]{\textsuperscript{35}} \\
\cline{2-6}
& GetTheScoop \cite{JiCu2014} & \textit{Static} & SVD of the graph’s adjacency matrix & Lockstep behaviors & WEIBO\hyperref[link16]{\textsuperscript{16}}, Synthetic data\\
\cline{2-6}
& CMD \cite{SuXi2007} & \textit{Dyn.} & Sparse Low rank decomposition (Compact Matrix Decomposition (CMD)) & Abnormal host (malicious network activities, misconfiguration errors, port scanning, DoS, DDoS) & Network flow, DBLP\hyperref[link3]{\textsuperscript{3}} \\
\cline{2-6}
& MAF \cite{GuMa2011} & \textit{Dyn.} & Performs Canonical Polyadic (CP) decomposition & Detect dense substructure patterns & LBNL\hyperref[link38]{\textsuperscript{38}}, RTW\hyperref[link37]{\textsuperscript{37}}, BDGP\hyperref[link36]{\textsuperscript{36}}\\
\cline{2-6}
& TensorSplat \cite{KoPa2012} & \textit{Dyn.} & Uses the "PARAFAC" decomposition method & Spotting micro-clusters, and changes & DBLP\hyperref[link3]{\textsuperscript{3}}, LBNL\hyperref[link38]{\textsuperscript{38}}, FACEBOOK\hyperref[link32]{\textsuperscript{32}}\\
\cline{2-6}
& DynAnom \cite{MaMa2013} & \textit{Dyn.} & (Sparsity and Low-rank) + Dynamic Anomalography & Abrupt unusual changes & Synthetic data, Internet-2\hyperref[link39]{\textsuperscript{39}} \\ 
\cline{2-6}
& MTHL \cite{TeLi2017} & \textit{Dyn.} & Multi-view Time-Series Hypersphere Learning (MTHL) + Support Vector Data Description (SVDD) & Abnormal events (civil unrest using social media, crowd activities or emergencies in cities, network intrusion or network failures) & Synthetic data, NYC taxi\hyperref[link6]{\textsuperscript{6}}, Twitter data \\
\cline{2-6}
& TMF \cite{YuAgWa2017} & \textit{Dyn.} & Temporal Matrix Factorization Model (TMFM) & Temporal anomalies and events & UCIMessages\hyperref[link9]{\textsuperscript{9}}, Digg\hyperref[link10]{\textsuperscript{10}}, EPINIONS\hyperref[link18]{\textsuperscript{18}}, Infectious\hyperref[link40]{\textsuperscript{40}}, arXiv hep-th\hyperref[link11]{\textsuperscript{11}}, DBLP\hyperref[link3]{\textsuperscript{3}} \\
\hline
\end{tabular}
\end{table*}

\subsection{Distance/Similarity Based Techniques}
Distance/Similarity-based techniques propose certain change measures of graph structures and use those measures to detect anomalies. These measures include embeddings, sketchings, PageRank scores, etc., of nodes, edges, and/or (sub)graphs. The following are the details of some of the anomaly detection methods that use Distance/Similarity-based techniques. 

In software behavior graphs, SoftBug \cite{LiYa2005} uses graph mining and SVM classification techniques to detect suspicious regions of non-crashing bugs. Here, the summary of software behavior graphs is represented by the frequent (closed) sub-graphs using the graph mining algorithm, $CloseMine$. Then, an SVM classifier is trained by the similarity measure between the vector representations to detect incorrect program runs. Outlier detection problems have mostly been studied in the context of attributes of nodes. That is, they use either density-based or distanced-based techniques to detect outliers. A density-based method compares the density around the nodes and the densities around its neighbors to detect outliers. The node is defined as a distance outlier if a node is further away than the minimum distance in multi-dimensional space. Most outlier detection techniques take more time to detect outliers if the data is very large or dynamic. So, LOADED \cite{GhOt2004} (Link-based Outlier and Anomaly Detection in Evolving Data sets) used a one-pass algorithm that can trade off accuracy with detection time. It is an outlier detection technique that uses link strength along with correlation statistics to model the similarity between data nodes having both categorical and continuous attributes. In other words, it computes the similarity score based on the combination of link strength for continuous attributes and the correlation statistics for categorical attributes. Nodes that violate these dependencies are considered outliers. NFAD \cite{SuQu2005} modeled real-world networks like authors vs. conferences of a publication network, traders vs. stocks of trading networks, etc., using bipartite graphs. It uses a random walk with a restart model to represent neighborhood information and detect outlier nodes. OutRank \cite{MoTa2008} uses the random walks method to detect outliers. It represents graph data based on the node similarity and number of common neighbors between nodes. The main importance of the random walks method is that it can detect not only scattered outliers but also small clusters of outliers. These small clusters of outliers can not be detected through previous distance-based and density-based outlier detection methods like K-dist and LOF, respectively. OutRank uses a cosine distance measure, whereas LOF \cite{BrKr2000} and K-dist \cite{JiTu2001} use Euclidean distance and (1 – cosine) distance to detect outliers.

The outlier detection problems are inherently challenging with the high volume of network streams represented as a dynamic graph. Streaming networks also make outlier detection methods computationally expensive. To address the outlier detection problems in streaming graphs, GOutlier \cite{AgZh2011} proposes an intricate structural analysis technique. It uses a probabilistic algorithm to maintain the structural summaries of underlying graph streams. In other words, it designs a structural reservoir sampling method to compress the structural information of graph streams. Besides detecting outlier nodes, similarity/distance-based methods are also used to discover unusual links in the graph networks. EbHo \cite{EbHo2007} uses a probabilistic and maximum partial substructure method to detect anomalous nodes and links. Their work uses the Minimum Description Length (MDL) to represent important substructures of a graph. While computing similarity, it examines the extensions of normal substructures with the lowest probability. Y{\scriptsize AGADA} \cite{DaLi2011} detects anomalies in labeled graphs using structural data and numeric attributes. Unlike EbHo \cite{EbHo2007}, it does not assume the prior distribution of numeric values. S{\scriptsize EDAN}S{\scriptsize POT} \cite{EsFa2018} and NETWALK \cite{YuCh2018} are edge streaming methods to detect anomalies in graph data. S{\scriptsize EDAN}S{\scriptsize POT} uses the proximity (distance) between a new edge and a set of sampled edges to detect bursts of activity. NETWALK proposes a graph embedding model to detect anomalies by performing a random walk on dynamic graphs.

CatchSYNC \cite{JiCu2016} is an anomaly detection method that represents each node in a features vector based on the node's behavior patterns, such as  degree values (in and out degrees) and HITS score (Hubness and authoritativeness). This approach computes the behavior patterns of the source node $u$ based on $synchronicity$: how a synchronized source's targets are in in-degree vs. authoritativeness feature space; and $normality$: how a normal source's targets are compared to other data. These two measures map the source's target nodes in the feature space, and the distance(similarity) between target nodes helps to detect a suspicious source node $u$. StreamSpot \cite{MaMi2016}, SpotLight \cite{EsFaGu2018}, GODIT \cite{PaMu2019}, and SnapSketch \cite{PaEb2020} are some graph sketching approaches to detect anomalies in streaming graphs. SpotLight sketches the graph into a $K$-dimensional features vector by randomly selecting $K$ subgraphs by sampling each source and destination node with probabilities. StreamSpot, GODIT, and SnapSketch use the shingling technique to sketch the graph. These approaches create a bag of k-shingles from a graph stream and construct shingle-frequency vectors to discover the similarity between the graphs.

\begin{table*}
\caption{Distance/Similarity Based Techniques}
\begin{tabular}{|p{1.5cm}|p{2.2cm}|p{0.8cm}|>{\baselineskip=7pt}p{4.1cm}|>{\baselineskip=7pt}p{3.3 cm}|>{\baselineskip=7pt}p{3.5cm}|}
\hline
\textbf{Anomalous Component} & \textbf{Papers} & 
\textbf{Graph Types} & \textbf{Techniques} & \textbf{Anomalies}& \textbf{Datasets}\\
\hline
\textit{Node} 
& SoftBug \cite{LiYa2005} & \textit{Static} & Closed Frequent Sub-graph Mining & Software Behavior Graphs (non-crashing bugs) & Buggy Code\\ 
\cline{2-6}
& NFAD \cite{SuQu2005} &  \textit{Static} & Random walks with restarts + graph partitioning & Outlier nodes P2P systems, Citation networks, etc. & Conference-Author(CA), Author-Paper(AP), IMDB\\
\cline{2-6}
& OutRank \cite{MoTa2008} & \textit{Static} & Similarity measure on Random Walks & Outliers Detection & 2D-Data, Austra,
Zoo, Diabetic, Led, Lymph, Pima, Vehicle, Optical, KDD-99 \hyperref[link13]{\textsuperscript{13}} \\
\cline{2-6}
& LOADED \cite{GhOt2004} &  \textit{Dyn.} & Link-based approach with correlation statistics & Outliers like Network Intrusions Detection & KDDCup\hyperref[link13]{\textsuperscript{13}}, Adult Database, Congressional Votes\\ 
\cline{2-6}
& GOutlier \cite{AgZh2011} & \textit{Dyn.} & Reservoir sampling method to maintain structural summaries & Structural Outlier Detection & DBLP\hyperref[link3]{\textsuperscript{3}}, IMDB\hyperref[link12]{\textsuperscript{12}} \\ 
\hline

\textit{Edge} 
& EbHo \cite{EbHo2007} &  \textit{Static} & MDL with other probabilistic measures & Fraud Detection, Intrusion Detection & Synthetic data, Cargo Network, KDDCup-1999\hyperref[link13]{\textsuperscript{13}}\\
\cline{2-6}
& NetWalk \cite{YuCh2018} & \textit{Dyn.} & Random walks + Deep auto-encoder neural network + reservoir sampling & Cyber-bullying, Terrorist attack planning, Advanced persistent threat (APT) & UCI Messages\hyperref[link9]{\textsuperscript{9}}, arXiv hep-th\hyperref[link11]{\textsuperscript{11}}, Digg\hyperref[link10]{\textsuperscript{10}}, DBLP\hyperref[link3]{\textsuperscript{3}}\\
\cline{2-6}
& SedanSpot \cite{EsFa2018} & \textit{Dyn.} & Edge Sampling and Marginal proximity distance & Port-scan, DoS, Scams (malicious entities attacking many victims), Occasions (holidays producing a burst traffic) & DARPA\hyperref[link4]{\textsuperscript{4}}, DBLP\hyperref[link3]{\textsuperscript{3}}, ENRON\hyperref[link1]{\textsuperscript{1}}\\
\hline

\textit{(Sub)Graph} 
& Y{\scriptsize AGADA} \cite{DaLi2011} &  \textit{Static} & Consider both structural data and numeric attributes & suspicious events - DoS & Access Control system\\
\cline{2-6}
& (Anti)Social \cite{DiKa2012} & \textit{Static} & Used community detection, cut-vertices, and local graph structure techniques & Intrusion detection & Net flow data from European ISP\\
\cline{2-6}
& GOutRank \cite{MuSa2013} & \textit{Static} & Used degree of deviation in both graph and attribute properties & Fraud detection, Network intrusion analysis & Amazon Co-purchase Network\hyperref[link46]{\textsuperscript{46}}\\
\cline{2-6}
& CatchSYNC \cite{JiCu2016} & \textit{Static} & Measure synchronicity and normality (degree values and HITS score) & Lockstep behaviors, DDoS attacks & TWITTERSG, WEIBO\hyperref[link16]{\textsuperscript{16}}\\
\cline{2-6}
& MetricFor \cite{HeEl2010}  & \textit{Dyn.} & Using metrics and analysis techniques (such as ego-net analysis) & Rare events (“elbow” pattern, prolonged spikes, and broken correlations) & ENTP, RMBT\hyperref[link47]{\textsuperscript{47}}, LBNL\hyperref[link38]{\textsuperscript{38}}\\
\cline{2-6}
& StreamSpot \cite{MaMo2016} & \textit{Dyn.} &  Shingling using Ordered k-hop Breadth First Traversal + Sketching (StreamHash)  & Advanced persistent threat (APT) detection & 
ALL (YouTube, Download, CNN, GMail, VGame)\hyperref[link45]{\textsuperscript{45}} \\ 
\cline{2-6}
& DenseAlert \cite{ShHo2017} &  \textit{Dyn.} & DenseAlert (an algo. for detecting suddenly emerging dense subtensors) & DoS attacks, Lockstep behaviors & (Yelp, Android, YahooM,  KoWiki, EnWiki, Youtube, SMS)\hyperref[link44]{\textsuperscript{44}}, DARPA\hyperref[link4]{\textsuperscript{4}} \\
\cline{2-6}
& SpotLight \cite{EsFaGu2018} & \textit{Dyn.} & SpotLight graph sketching (K specific subgraphs chosen based on node sampling probabilities) & DoS attack, Lockstep behavior, Unusual events & DARPA\hyperref[link4]{\textsuperscript{4}}, Enron\hyperref[link1]{\textsuperscript{1}}, NYCTaxi\hyperref[link6]{\textsuperscript{6}} \\
\cline{2-6}
& A{\scriptsize NOM}R{\scriptsize ANK} \cite{YoHo2019} & \textit{Dyn.} & \textit{AnomalyS} (suspicious changes to the structure) and \textit{AnomalyW} (anomalous changes in the weight) & Link spam, Follower boosting,  DDoS attacks & ENRON\hyperref[link1]{\textsuperscript{1}}, DARPA\hyperref[link4]{\textsuperscript{4}} \\
\cline{2-6}
& GODIT \cite{PaMu2019} & \textit{Dyn.} & Shingling using Biased Random Walk + Sketching using Discriminative Shingle & DoS attack on  IoT-equipped smart home & UNSW-NB15\hyperref[link7]{\textsuperscript{7}} \\
\cline{2-6}
& SnapSketch \cite{PaEb2020} & \textit{Dyn.} & Shingling using Biased Random Walk + Sketching (Simplified Hashing) & Port scan, DoS attack, Lockstep behavior, Unusual events & UNSW-NB15\hyperref[link7]{\textsuperscript{7}}, DARPA\hyperref[link4]{\textsuperscript{4}}\\
\hline
\end{tabular}
\end{table*}

\subsection{Graph Neural Networks Techniques}
The techniques mentioned earlier (such as probabilistic/statistical methods, matrix/tensor factorization methods, and distance/similarity-based methods) have been the most common approaches for detecting anomalies in graph-structured data. Additionally, some approaches have used traditional machine-learning approaches like random forests, decision trees, and k-nearest neighbors to detect anomalies in graph data. However, recently, Graph Neural Networks (GNNs) like Graph Convolutional Networks (GCNs), Graph Attention Networks (GATs), and Graph AutoEncoder (GAE) techniques have become more popular approaches as they can represent the complex relationships present in graph data and detect anomalies.

Graph learning techniques like a GCN uses graph convolutions, and a GAT uses graph neural networks to learn graph-structured data for anomaly detection. Like traditional convolutional networks, a GCN also learns graph data by aggregating and transforming node features through multiple layers of convolutions. Then, the result of the GCN is used to predict anomalies. Likewise, a GAT learns to encode graph data into a lower dimension while preserving important features and relationships between nodes using attention mechanisms (i.e., by aggregating node features and updating node representations through multiple layers). Then, the final node representations are used to predict anomalies. A GAE encodes the input graph data into a low-dimensional vector space representation while also decoding to reconstruct the original input graph from the vector space representation. The encoded vector representation preserves the essential features and patterns of the normal behavior within the graph. The significant differences between the original and reconstructed graph data in the vector space are considered anomalies.

\begin{table*}
\caption{Graph Neural Networks Techniques}
\begin{tabular}{|p{1.5cm}|p{2.2cm}|p{0.8cm}|>{\baselineskip=7pt}p{4.1cm}|>{\baselineskip=7pt}p{3 cm}|>{\baselineskip=7pt}p{3.8cm}|}
\hline
\textbf{Anomalous Component} & \textbf{Papers} & 
\textbf{Graph Types} & \textbf{Techniques} & \textbf{Anomalies}& \textbf{Datasets}\\
\hline
\textit{Node} 
& DOMINANT \cite{Dominant2019} & \textit{Static} & GCN-based autoencoder on attributed networks & Detecting anomalies in the different networks & BlogCatalog\hyperref[link52]{\textsuperscript{52}}, Flickr\hyperref[link32]{\textsuperscript{32}}, ACM\hyperref[link24]{\textsuperscript{24}} \\
\cline{2-6}
& SpecAE \cite{SpecAE2019} & \textit{Static} & GCN-based autoencoder + Graph De-convolutional Networks & Detecting anomalies citation networks & CORA\hyperref[link19]{\textsuperscript{19}}, PubMed\hyperref[link20]{\textsuperscript{20}}\\
\cline{2-6}
& FdGars \cite{FdGars2019}  & \textit{Static} & Graph Convolutional Neural Networks & Fake review detection & Data from Tencent’s Venus Computation Platform\\
\cline{2-6}
&GCNSI \cite{GCNSI2019}& \textit{Static} & Label based Propagation Source Identification (LPSI) + GCN & Rumor source detection & (Karate, Dolphin, Jazz)\hyperref[link48]{\textsuperscript{48}} \\
\cline{2-6}
&SemiGNN \cite{SemiGNN2019} &\textit{Static}& Hierarchical attention structure in GNN & Financial Fraud Detection & ALIPAY\\
\cline{2-6}
&GCNwithMRF \cite{GCNwithMRF2020} &\textit{Static}& GCN + Markov Random Fields (MRFs) & Spam detection& Twitter Social Honeypot\hyperref[link49]{\textsuperscript{49}}, Twitter 1KS-10KN\\ 
\cline{2-6}
&(D)TPC-GCN\cite{TPCGCN2020}  &\textit{Static}& Topic-Post-Comment (TPC) Graph + GCN & Controversial post detection & WEIBO\hyperref[link16]{\textsuperscript{16}} and Reddit\hyperref[link28]{\textsuperscript{28}}\\
\cline{2-6}
& ALARM \cite{ALARM2020}  & \textit{Static} & Attributed Networks with Graph Convolutional Neural Networks & Detecting outliers & WebKB\hyperref[link22]{\textsuperscript{22}}, Cora\hyperref[link19]{\textsuperscript{19}}, CiteSeer\hyperref[link21]{\textsuperscript{21}}, PubMed\hyperref[link20]{\textsuperscript{20}}\\ 
\cline{2-6}
& GraphRfi \cite{GraphRfi2020} &\textit{Static} & GCN + Neural Random Forest (NRF) & Fake review detection & Yelp\hyperref[link2]{\textsuperscript{2}}, Movies\&Tv\hyperref[link50]{\textsuperscript{50}} \\
\cline{2-6}
& GraphConsis \cite{GAN2020} & \textit{Static} & Handle inconsistency problems of GNN  & Fraud detection, Fake review detection & Cora\hyperref[link19]{\textsuperscript{19}}, PPI\hyperref[link51]{\textsuperscript{51}}, Reddit\hyperref[link28]{\textsuperscript{28}}, Yelp\hyperref[link2]{\textsuperscript{2}}\\
\cline{2-6}
&CoLA \cite{CoLA2022} & \textit{Static} & GCN-based contrastive learning model &Anomaly detection in the different networks &BlogCatalog\hyperref[link52]{\textsuperscript{52}}, Flickr\hyperref[link53]{\textsuperscript{53}}, Cora\hyperref[link19]{\textsuperscript{19}}, CiteSeer\hyperref[link21]{\textsuperscript{21}}, PubMed\hyperref[link20]{\textsuperscript{20}}, ACM\hyperref[link24]{\textsuperscript{24}} and ogbn-arxiv \hyperref[link23]{\textsuperscript{23}}\\
\cline{2-6}
&GraphCAD\cite{GCCAD2023}& \textit{Static}& Context-aware GNN encoder& Detecting fake review, financial frauds, fake users & (AMiner, MAS, Alpha, Yelp)\hyperref[link54]{\textsuperscript{54}}\\
\cline{2-6}
& GEM \cite{GEM2020} & \textit{Dyn.} & GCN or GCN with attention & Detecting malicious accounts& Alipay\\
\cline{2-6}
& OCAN \cite{OCAN2019} & \textit{Dyn.} & LSTM-Autoencoder + GAN model & Fraud detection & UMDWikipedia\hyperref[link55]{\textsuperscript{55}}\\
\cline{2-6}
& GCAN \cite{GCAN2020} & \textit{Dyn.} & CNN + GRU + GCN &Fake news detection & Twitter15, Twitter16\\ 
\cline{2-6}
&GDN \cite{GDN2021} &\textit{Dyn.}& Garph Structure Learning + GAT &Anomalies detection on CPS System & (Secure Water Treatment, Water Distribution)\hyperref[link56]{\textsuperscript{56}}\\
\cline{2-6}
&TADDY \cite{TADDY2023} & \textit{Dyn.} & Dynamic Graph Transformer & Fake review, fake trading & UCIMessages\hyperref[link9]{\textsuperscript{9}}, Digg\hyperref[link10]{\textsuperscript{10}}, Email-DNC\hyperref[link57]{\textsuperscript{57}}, Bitcoin-Alpha\hyperref[link58]{\textsuperscript{58}}, Bitcoin-OTC\hyperref[link59]{\textsuperscript{59}}, AS-Topology\hyperref[link60]{\textsuperscript{60}}\\
\hline

\textit{Edge} 
& AANE \cite{AANE2020} & \textit{Static} & Graph Convolution based GAE & Anomalous links detection & (Disney, Books)\hyperref[link61]{\textsuperscript{61}}, ENRON\hyperref[link1]{\textsuperscript{1}}, BlogCatalog\hyperref[link52]{\textsuperscript{52}}, Flickr\hyperref[link32]{\textsuperscript{32}}, ACM\hyperref[link24]{\textsuperscript{24}}\\
\cline{2-6}
& SubGNN \cite{SubGNN2021} & \textit{Static} &  Relational Graph Isomorphism Network (R-GIN) + GNN & Fraud detection & Amazon, Yelp\hyperref[link2]{\textsuperscript{2}}, Taobao\\
\cline{2-6}
& eFraudCom \cite{eFraudCom2022} & \textit{Static} & Graph Convolutional based GAE & Fraud detection & Taobao, MOOC\hyperref[link62]{\textsuperscript{62}}, Bitcoin-Alpha\hyperref[link58]{\textsuperscript{58}}\\
\cline{2-6}
& AddGraph \cite{LiZh2019} & \textit{Dyn.} & GCN + GRU & Detection of fake pages, fake recommendations & UCI Message\hyperref[link9]{\textsuperscript{9}}, Digg\hyperref[link10]{\textsuperscript{10}}\\
\cline{2-6}
&H-VGRAE \cite{HVGRAE2020} & \textit{Dyn.} & Variational Graph Autoencoder + RNN & Anomalous link on various networks & UCI Message\hyperref[link9]{\textsuperscript{9}}, arXiv HEP-TH\hyperref[link11]{\textsuperscript{11}}, Social evolution\hyperref[link63]{\textsuperscript{63}}, Github\hyperref[link64]{\textsuperscript{64}}\\
\cline{2-6}
&DynAD \cite{DynAD2020} & \textit{Dyn.} & GCN + GRU & Detection of fake pages, fake recommendations & UCIMessages\hyperref[link9]{\textsuperscript{9}}, arXiv hep-th\hyperref[link11]{\textsuperscript{11}}, Digg\hyperref[link10]{\textsuperscript{10}}\\
\cline{2-6}
&Hierarchical-GCN \cite{HGCN2020} & \textit{Dyn.} & GCN + GRU & Discovering anomalies in the different networks& UCI Message\hyperref[link9]{\textsuperscript{9}}, ENRON\hyperref[link1]{\textsuperscript{1}}, Facebook\hyperref[link65]{\textsuperscript{65}}, Math\hyperref[link66]{\textsuperscript{66}}\\
\cline{2-6}
& Bi-GCN \cite{BiGCN2020} & \textit{Dyn.}  & Top-Down GCN and Bottom-UP GCN & Rumor Detection & WEIBO\hyperref[link16]{\textsuperscript{16}}, Twitter15, and Twitter16 \\
\cline{2-6}
& StrGNN \cite{StrGNN2021} & \textit{Dyn.} & GCN + GRU & Detecting anomalies in the different & Bitcoin-alpha\hyperref[link58]{\textsuperscript{58}} and Bitcoin-otc\hyperref[link59]{\textsuperscript{59}}, Digg\hyperref[link10]{\textsuperscript{10}}, Email, Topology, UCIMessage\hyperref[link9]{\textsuperscript{9}}\\
\hline

\textit{(Sub)Graph}
&MatchGNet\cite{MatchGNet2019}&\textit{Static}& Invariant Graph Modeling (IGM) + Hierarchical Attentional GAE & Fake program detection& Enterprise network data\\
\cline{2-6}
& mHGNN \cite{mHGNN2020} & \textit{Static} & Attributed Heterogeneous Information Network (AHIN) + Metagraph Aggregated Heterogeneous GNN & Illicit Traded Product Identification & Hack Forums\\
\cline{2-6}
& GLocalKD \cite{GLocalKD2022} & \textit{Static}  & Use two GNNs and joint random knowledge distillation of graph and node representations & Drug discovery, money laundering, molecular synthesis, etc & (PROTEINS, ENZYMES, AIDS, DHFR, and other more 12 datasets)\hyperref[link67]{\textsuperscript{67}}\\
\cline{2-6}
& OCGTL \cite{OCGTL2022} & \textit{Static}  & One-class classifcation (OCC) + GNN & Drug discovery, money laundering, molecular synthesis, etc & (DD, PROTEINS, ENZYMES, NCI1, AIDS, and Mutagenicity, IMDB, REDDIT)\hyperref[link67]{\textsuperscript{67}}\\
\cline{2-6}
& HO-GAT \cite{HoGAT2023} & \textit{Static} & Motif instance representations are integrated with GAT & Web spam detection, system fraud detection, network intrusion detection & Scholat\hyperref[link68]{\textsuperscript{68}}, AMiner\hyperref[link69]{\textsuperscript{69}}, and four WebKB\hyperref[link22]{\textsuperscript{22}} \\
\cline{2-6}
& OCGIN \cite{OCGIN2023} & \textit{Static} & GNN + Graph Isomorphism Network (GIN)& Drug discovery, money laundering, molecular synthesis, etc.& (DD, PROTEINS, NCI, and IMDB)\hyperref[link67]{\textsuperscript{67}}\\
\cline{2-6}
&ST-GCAE \cite{ST-GCAE2020} & \textit{Dyn.} & Spatio-Temporal GAE & Video anomaly detection & ShanghaiTech Campus dataset,  NTU-RGB+D\hyperref[link70]{\textsuperscript{70}}, Kinetics-250 \\
\cline{2-6}
& TGBULLY \cite{TGBULLY2021}& \textit{Dyn.} & GRU + GAT & Cyberbullying detection in social networks & Instagram, Vine \\
\hline

\end{tabular}
\end{table*}

In recent years, there has been extensive graph-based anomaly detection research using Graph Neural Networks (GNNs) techniques. DOMINANT \cite{Dominant2019} states that traditional anomaly detection techniques face challenges:
\begin{itemize}
\item \textbf{Network sparsity}: real-world networks could be very sparse and ego-net or community analysis might be difficult in such sparse networks
\item \textbf{Data nonlinearity}: the node interactions might be non-linear while traditional methods assume to be linear
\item \textbf{Complex interactions}: traditional methods often struggle with the complex interconnected nature of graph data.
\end{itemize}

To address those challenges, DOMINANT presents a GCN-based autoencoder (GAE) to detect anomalies on attributed networks. Likewise, to detect anomalies on attributed networks, SpecAE \cite{SpecAE2019} uses a GCN-based autoencoder to learn node representations and Graph De-convolution Networks to reconstruct nodal attributes based on topological relations. ALARM \cite{ALARM2020} and GraphCAD \cite{CoLA2022} are other GCN-based anomaly detection approaches on attributed networks. In FdGars \cite{FdGars2019} work, the authors use the GCN to detect fake reviews on e-commerce networks and claim that their approach is more effective and scalable than the normal learning methods like Logistic Regression, Random Forest, etc. SemiGNN \cite{SemiGNN2019} notes that there are very few labeled data compared to unlabeled data for fraud detection. They use a semi-supervised Graph Attention Network (GAT) to detect fraud detection in financial networks, and compare their results with standard machine learning models along with GCN and GAT models. Like SemiGNN \cite{SemiGNN2019} that uses labeled and unlabeled information, GCNSI \cite{GCNSI2019} propagates the node's label information in the network and passes them through the GCN model to predict the rumor sources on social networks. In a similar way, GCNwithMRF \cite{GCNwithMRF2020} is another semi-supervised GCN model that uses a Markov Random Field (MRF) as a Recurrent Neural Network (RNN) to capture the neighbors' influences on a user's identity and detect spam that follows a large number of users. (D)TPC-GCN\cite{TPCGCN2020} initially generates the Topic-Post-Comment relationships (semantic and structural) graph and passes through the GCN model for post-level controversy detection on social networks. GraphConsis \cite{GAN2020} is another work that considers various inconsistencies like context inconsistency, feature inconsistency, and relation inconsistency along with a GCN to detect fraud in social networks. In GraphRfi \cite{GraphRfi2020}, the authors state that the current recommendation systems are unaware of fake reviews and use such misinformation in their learning model. So, their work proposes learning a framework with a GCN and Neural Random Forest to detect fake reviews in e-commerce data. Likewise, GraphCAD \cite{CoLA2022} and GraphCAD \cite{GCCAD2023} are other papers that use a contrastive learning (context-aware) GNN to detect fake reviews or ratings in e-commerce or social networks.

\cite{OCAN2019, GCAN2020, GDN2021} use types of RNNs to detect node anomalies in dynamic networks. OCAN \cite{OCAN2019} uses an LSTM-Autoencoder and Generative Adversarial Networks (GAN) to detect fraud in online social networks. Likewise, GCAN \cite{GCAN2020} is another work that uses a Convolutional and Gated Recurrent Unit (GRU) to learn a representation of retweet propagation and a GCN to learn a representation of user interactions to detect fake news in social networks. In GDN \cite{GDN2021}, the authors use Graph Structure Learning to learn relationships and Graph Attention-based Forecasting to detect anomalies in Cyber-Physical Systems (CPS). They compare their results with an Autoencoder and various LSTM models. Whereas, TADDY \cite{TADDY2023} recently used the graph transformer framework to model spatial and temporal information and detect anomalies in the dynamic graph networks. 

\cite{AANE2020, SubGNN2021, eFraudCom2022, LiZh2019, HVGRAE2020, DynAD2020, HGCN2020, BiGCN2020, StrGNN2021} are some other research papers that use a GCN to detect anomalous edges in graph networks. AANE \cite{AANE2020} and eFraudCom \cite{eFraudCom2022} use a Graph Convolution based GAE, and SubGNN \cite{SubGNN2021} uses a Relational Graph Isomorphism Network (R-GIN) and GNN for fraud detection in static graphs. On the other hand, AddGraph \cite{LiZh2019}, DynAD \cite{DynAD2020}, Hierarchical-GCN \cite{HGCN2020}, and StrGNN \cite{StrGNN2021} use a GCN and a GRU to preserve the structural and temporal aspects, and detect anomalous edges in dynamic networks. StrGNN \cite{StrGNN2021} proposes an end-to-end anomalous edge detection using the structural temporal graph neural network model whereas Addgraph \cite{LiZh2019} uses an extended temporal GCN with an attention model. However, Hierarchical-GCN \cite{HGCN2020} claims that StrGNN \cite{StrGNN2021} and Addgraph \cite{LiZh2019} do not consider the hierarchical information, so they combine the hierarchical data evaluation with temporal information to detect anomalies in dynamic networks. Likewise, HVGRAE \cite{HVGRAE2020} combines an RNN with a Variational Autoencoder (VE), and extends them to the graph domain by extracting the spatial aspects using a GNN. The authors state that their proposed approach considers the complex spatial-temporal (ST) patterns that lead to detect sensitive local and short-term changes that could not be detected by approaches like AddGraph \cite{LiZh2019}. In the Bi-GCN \cite{BiGCN2020} paper, they conjecture that different RNN approaches only consider the temporal-structural features, and only focus on the sequential propagation of rumors without considering the influences of rumor dispersion. They then present a Bi-directional GCN to detect rumors in social networks.

ST-GCAE \cite{ST-GCAE2020} and TGBULLY \cite{TGBULLY2021} are other approaches that also consider the temporal aspects using the GCN technique. However, these papers only detect (sub)graph anomalies in dynamic networks. ST-GCAE \cite{ST-GCAE2020} uses the spatiotemporal features to detect posture anomalies in the video, whereas TGBULLY \cite{TGBULLY2021} uses semantic context and temporal graph interactions to detect cyberbullying. While there is very little GCN-based research happening to detect anomalies in (sub)graph streams, there has been some research \cite{MatchGNet2019, mHGNN2020, GLocalKD2022, OCGTL2022, HoGAT2023, OCGIN2023} in detecting (sub)graph anomalies in static networks. MatchGNet \cite{MatchGNet2019} and HO-GAT \cite{HoGAT2023} use Graph Attention Networks (GAT), and \cite{mHGNN2020, GLocalKD2022, OCGTL2022, OCGIN2023} use a graph information technique along with the GCN model to detect anomalous (sub)graphs. 
\section{Type of Anomaly}\label{ta}
Recent research has provided valuable insight into the task of detecting anomalies in large data sources that can be represented in a graph format. The results of these efforts have provided analytical tools for discovering anomalies in a wide variety of application domains, including social network analysis, fraud detection, intrusion detection, etc. Most anomaly detection techniques use various network data sets (i.e., real and synthetic) to detect different types of anomalies, such as Outlier detection, Change detection, Lockstep behaviors, DoS attacks, DDoS attacks, Port scanning, etc. In this section, we broadly discuss the different types of anomalies: Outliers, Dense (Group) Anomalies, Sudden Anomalies, and Gradual Anomalies. 

One could consider all types of anomalies as outliers. But, for our classification, we refer to outliers as meaning small numbers of entities that are deviating from a pattern that is represented by the majority of normal entities. \textit{Outlier detection} techniques are mostly used to detect this kind of anomaly. \textit{Dense (Group) anomalies}, such as Lockstep behaviors, DoS attacks, Port scanning, etc., are a type of anomaly that harm the networks in the group. A sudden change of community, sudden appearance/disappearance, and sudden change of distribution, etc., are what we are referring to as \textit{sudden anomalies}. Most anomaly detection techniques discussed in this survey fall into Sudden and/or Group Anomalies. DTA/STA \cite{SuTa2006}, CoreScope \cite{ShEl2018}, NrMF \cite{ToLi2012}, CatchSYNC \cite{JiCu2016} are a few of the static graph approaches for detecting group anomalies. CopyCatch \cite{BeXu2013}, MIDAS \cite{BhHo2020}, Isconna \cite{LiSi2021}, PENminer \cite{BeZh2020}, TF-IGF \cite{LaEb2021}, MTHL \cite{TeLi2017}, SedanSpot \cite{EsFa2018}, SnapSketch \cite{PaEb2020}, SpotLight \cite{EsFaGu2018} etc., are some of the dynamic approaches that detect sudden group anomalies. Likewise, F-FADE \cite{ChLi2021} detects sudden group anomalies and sudden community change. Again, most of the research in graph-based anomaly detection is focused on detecting sudden/group anomalies. However, a final type of anomaly, gradual anomalies, refers to stealthy anomalies in which attackers gradually enter the network to try and remain undetectable. However, there is very little research going on to detect these gradual anomalies. Ideally, one would like an approach that can handle sudden and gradual anomalies. 
\section{Applications}\label{app}
Graph-based anomaly detection methods analyze the topology, connections, and features of a graph to identify unusual behaviors that deviate significantly from normal behavior. This method is particularly effective in detecting anomalies in complex systems such as social networks, communication networks, e-commerce networks, road networks, financial networks, etc. The following are some of the applications of graph-based anomaly detection.   

\subsection{Social Networks} 
Graph-based anomaly detection techniques in social networks focus on network structure, user-user interactions, and content to detect unusual behaviors. For instance, lockstep patterns such as sudden page likes or friend requests, unusual post patterns, or accounts behaving more absurdly than their peers could represent spam activity or botnet attacks. Many graph based anomaly detection papers \cite{BhHo2020, BeZh2020, LaEb2021, BhLi2022, ShBe2014, JiCu2016, HoSh2017, ShEl2018, BeXu2013, ChNe2014, JiBe2015, ShHo2016, ShHoKi2017, JiCu2014, KoPa2012, TeLi2017, TeLi2017, MaMo2016, ShHo2017, EsFaGu2018, Dominant2019, FdGars2019, GCNSI2019, GCNwithMRF2020, TPCGCN2020, ALARM2020, GAN2020, CoLA2022, GCAN2020, TADDY2023, AANE2020, LaEb_2022, LiZh2019, HVGRAE2020, DynAD2020, HGCN2020, BiGCN2020, OCGTL2022, HoGAT2023, TGBULLY2021} are applied to social networks. Among them, most papers detect sudden and bursty activities like lockstep behaviors, DoS or DDoS attacks on social media, and unusual events such as disease outbreaks, and terrorist attacks through posts on social media. Anomaly detection on social networks not only maintains integrity and security but also protects users from scams and fake information. The publicly available social network datasets are TwitterSecurity\hyperref[link14]{\textsuperscript{14}}, TwitterWorldCup\hyperref[link17]{\textsuperscript{17}}, Reddit\hyperref[link28]{\textsuperscript{28}}, Stackoverflow\hyperref[link29]{\textsuperscript{29}}, TWITTER\hyperref[link15]{\textsuperscript{15}}, WEIBO\hyperref[link16]{\textsuperscript{16}}, (Friendster, Orkut, Flickr, YouTube, Catster, Twitter)\hyperref[link32]{\textsuperscript{32}}, FACEBOOK\hyperref[link32]{\textsuperscript{32}}, Digg\hyperref[link10]{\textsuperscript{10}}, EPINIONS\hyperref[link18]{\textsuperscript{18}}, Bitcoin-Alpha\hyperref[link58]{\textsuperscript{58}}, Bitcoin-OTC\hyperref[link59]{\textsuperscript{59}}, and Youtube\hyperref[link44]{\textsuperscript{44}} etc.

\subsection{Computer Networks}
In computer networks, a graph-based anomaly detection model uses the network's topological structure, device connections, and traffic data to detect unusual activities. These activities include sudden attacks like DoS or DDoS attacks, port scans, hacking or intrusion attempts, and data leaks or breaches, etc.  \cite{BhHo2020, BeZh2020, LaEb2021, LiSi2021, BhLi2022, BhJa2021, MoTa2008, EbHo2007, EsFa2018, ShHo2017, EsFaGu2018, YoHo2019, PaEb2020} are some anomaly detection papers to detect intrusion in datasets like DARPA\hyperref[link4]{\textsuperscript{4}}, CIC-IDS\hyperref[link25]{\textsuperscript{25}}, ISCX-IDS\hyperref[link25]{\textsuperscript{25}}, and KDDCUP99\hyperref[link13]{\textsuperscript{13}}. Likewise, the graph-based anomaly detection papers \cite{LaEb2021, LiSi2021, BhLi2022} detect DoS or DDoS attacks in datasets like CTU-13\hyperref[link8]{\textsuperscript{8}} and CIC-DDoS\hyperref[link25]{\textsuperscript{25}}. Moreover, analyzing deviations from usual communication patterns could help identify advanced persistent threats (APTs) and other complex cyber-attacks.

\subsection{E-commerce Networks}
Many graph-based anomaly detection methods detect suspicious activities, such as unusual product recommendations/reviews, unusual purchase patterns, abnormal login attempts, suspicious payment patterns, etc., in e-commerce networks. Those methods mainly consider connections between users, products, and transactions to identify fraudulent actions. \cite{PaCh2007, ShBe2014, HoSh2017, MuSa2013, SubGNN2021, eFraudCom2022} are some graph-based anomaly detection techniques applied in e-commerce networks. Among them, \cite{PaCh2007} detects auction fraud like an unusual number of monetary loss reports, \cite{ShBe2014, HoSh2017} detect fake product reviews, and \cite{SubGNN2021, eFraudCom2022} detect fraudulent transactions. All of this ongoing research maintains trust and security for both customers and merchants in e-commerce networks. Amazon\hyperref[link30]{\textsuperscript{30}}, Amazon\hyperref[link71]{\textsuperscript{71}}, Amazon Co-purchase Network\hyperref[link46]{\textsuperscript{46}} are few publicly available e-commerce datasets.

\subsection{Financial Networks}
In financial networks, graph-based anomaly detection plays a crucial role in detecting unusual or suspicious behaviors such as unusual transaction patterns, irregular account behavior, unexplained withdrawals or transfers, unusual payment methods, etc. SemiGNN \cite{SemiGNN2019} and GEM \cite{GEM2020} use graph-based anomaly detection to identify malicious accounts in payment networks like Alipay. TADDY \cite{TADDY2023}, eFraudCom \cite{eFraudCom2022}, and StrGNN \cite{StrGNN2021}, detect fake trading on who-trusts-whom networks of bitcoin users (Bitcoin-Alpha\hyperref[link58]{\textsuperscript{58}}, Bitcoin-OTC\hyperref[link59]{\textsuperscript{59}}). Graph-based anomaly detection methods analyze the relationships between accounts and entities and complex financial transaction patterns to detect anomalies like money laundering, fraud, or account hijacking, etc. Early anomaly detection is especially important in financial institutions to defend against fraud on time and save from colossal financial damage.

\subsection{Citation and Collaboration Networks}
A lot of graph-based anomaly detection papers \cite{AkMc2010, RaHa2016, ShEl2018, YuAg2013, SuTa2006, Ch2004, ChLi2021, PrSe2009, SuXi2007, KoPa2012, YuAgWa2017, AgZh2011, YuCh2018, EsFa2018, EsFa2018, SpecAE2019, GAN2020, CoLA2022, HVGRAE2020, DynAD2020, GLocalKD2022, OCGTL2022, OCGIN2023} use citation or collaboration datasets. These papers mainly utilize publication patterns, citation behavior, and co-author relationships to detect anomalies like excessive self-citations, citation circles, etc. Graph-based anomaly detection methods maintain the quality and credibility of academic research networks by detecting citation fraud, research misconduct, or unethical publishing practices. There are many citation and collaboration datasets available: DBLP\hyperref[link3]{\textsuperscript{3}}, IMDB\hyperref[link12]{\textsuperscript{12}}, (LiveJournal, Patent)\hyperref[link32]{\textsuperscript{32}}, arXiv hep-th\hyperref[link11]{\textsuperscript{11}}, ACM\hyperref[link24]{\textsuperscript{24}}, CORA\hyperref[link19]{\textsuperscript{19}}, PubMed\hyperref[link20]{\textsuperscript{20}}, and CiteSeer\hyperref[link21]{\textsuperscript{21}}.

\subsection{Internet-of-Things (IoT) Networks}
There are a few papers \cite{LaEb2021, LiSi2021, BhLi2022, BhJa2021, PaMu2019, PaEb2020} that detect sudden and burst attacks like DoS or DDoS in an Internet-of-Things (IoT) dataset, such as UNSW-NB15\hyperref[link7]{\textsuperscript{7}}. \cite{LaMa_2022} another paper that detects DoS attacks on the Internet of Health Things (IoHT) networks. These graph-based anomaly detection papers consider the topological structure and interactions between IoT devices to detect sudden and abrupt increases in traffic, which might represent attacks like DoS or DDoS attacks, unauthorized access attempts, botnets, etc. Though all these papers are important for detecting sudden and abrupt anomalies, more research needs to be performed to detect other kinds of anomalies and ensure the integrity and security of IoT networks. 

\subsection{Road Networks}
In road networks, graph-based anomaly detection plays an important role in detecting irregularities such as traffic congestion, sudden spikes of accidents and incidents, road closures, route deviations, etc. \cite{BeZh2020, MoBo2013, TeLi2017, EsFaGu2018} are graph-based anomaly detection models which identify unusual events on road network datasets like NYCTaxi\hyperref[link6]{\textsuperscript{6}}, and Small and Large Traffic Networks\hyperref[link41]{\textsuperscript{41}}, by analyzing the road networks, traffic patterns, and data from sensors and GPS devices. These models help manage traffic and ensure the safety of commuters.

\subsection{Cyber-Physical Networks}
In Cyber-Physical networks, graph-based anomaly detection models are crucial to detect abnormalities like unexpected sensor device readings, unusual control commands, intrusion attempts, etc., and maintain the reliability and security of critical devices. These models focus on interactions between the control network system and the physical devices to make the cyber-physical network resilient. In other words, the model prevents system disruptions, sensor device failures, or cyber-attacks on targeted physical devices. GDN\cite{GDN2021} is the one paper that focuses on Cyber-Physical system, (Secure Water Treatment (SWaT) and Water Distribution (WADI))\hyperref[link56]{\textsuperscript{56}}.  

\hfill \break
\indent In summary, graph-based anomaly detection is an effective and powerful method to detect attacks, threats, irregularities, or unusual behaviors across various domains. EuEmail network\hyperref[link5]{\textsuperscript{5}}, ENRON\hyperref[link1]{\textsuperscript{1}}, and UCIMessages\hyperref[link9]{\textsuperscript{9}} are communication network datasets, (KoWiki, EnWiki)\hyperref[link44]{\textsuperscript{44}}, Wiki-vote\hyperref[link31]{\textsuperscript{31}}, and UMDWikipedia\hyperref[link55]{\textsuperscript{55}} are wikipedia network datasets, and (DD, PROTEINS, ENZYMES, NCI1, AIDS, and Mutagenicity)\hyperref[link67]{\textsuperscript{67}} are some molecules and bioinformatics datasets. Furthermore, Network Repository\hyperref[link72]{\textsuperscript{72}}, SNAP\hyperref[link73]{\textsuperscript{73}}, MIT Lincoln Laboratory\hyperref[link74]{\textsuperscript{74}}, infolab\hyperref[link75]{\textsuperscript{75}}, Konect\hyperref[link76]{\textsuperscript{76}}, Stratosphere Lab\hyperref[link77]{\textsuperscript{77}}, ANLAB\hyperref[link78]{\textsuperscript{78}}, Data Repository at ASU\hyperref[link79]{\textsuperscript{79}}, and CAIDA\hyperref[link80]{\textsuperscript{80}} are some of the major dataset repositories that contain a variety of network datasets on various domains.
\section{Challenges and Future Directions}\label{cfd}
In this section, we present the challenges faced by the current graph-based anomaly detection methods along with potential future research directions. The following are the details of the major challenges and future directions.  

\subsection{Dynamic Graphs}
Many real-world graphs are dynamic so it is crucial to consider the temporal aspects efficiently to capture anomalies. However, it is challenging to represent the evolving graph structure and identify anomalies in temporal patterns. Apart from the graphs being dynamic, the current graph data are also streaming in nature, where graphs evolve continuously, and there is a need to detect anomalies in near real-time so that a remedy can be applied as soon as possible and minimize the loss. Therefore, more efforts in extending graph-based anomaly detection research to handle dynamic (streaming) graph data are needed. 

\subsection{Graph Complexity} 
In recent years, the size of graphs has increased exponentially. The challenges of detecting anomalies on such graphs are also growing along with the graph size. With the increased graph size, the interconnections between various similar domains and the graph heterogeneity have led to an increase in the complexity of the graph. Thus, it is very challenging to represent such complex graphs and detect anomalies. Likewise, with the increase in graph size, the dimensionality to represent the graph also increases. Dealing with such high-dimensional graph data can be computationally very challenging. Most graph data, like social networks, citation networks, etc., are large and sparse. Identifying graph-based anomalies is challenging since very few relationships exist compared to the entire large network. In other words, anomaly detection techniques may struggle to preserve the proper structure of the graph. Since graph size is increasing exponentially, graph-based anomaly detection requires a scalable method to handle large-scale graph data. Thus, since graph complexities add a lot of challenges it is essential to consider these complexities when building the proper graph-based anomaly detection models. 

\subsection{New Application Areas} 
Many existing graph-based anomaly detection methods are applied to social networks, communication networks, e-commerce networks, citation and collaboration networks, etc. However, there is very little research going on in other domains like cyber-physical networks, IoT networks, healthcare networks, supply chain networks, etc. Graph-based anomaly detection method is important to identify sensor or controller failures and attacks on networks like IoT, smart grids, and industrial control systems. Detecting anomalies like natural calamities, pollution spikes, unusual weather patterns, etc., by analyzing the environmental monitoring systems would be a new domain for graph-based anomaly detection. Likewise, graph-based anomaly detection models can be applied to identify unusual events like pandemics, false insurance claims in healthcare networks, and artificial scarcity or unusual shipments in supply chain networks.

\subsection{Interpretability and Explainability} 
Many graph-based anomaly detection methods, like Graph Neural Networks (GNNs), are often black-box models that initially represent complex graph structures and then apply numerous weights on graph representations through multiple layers to detect anomalies. All these complex processes make GNN models harder to interpret and explain anomaly predictions. Additionally, some graph-based anomaly detection techniques use an embedding approach to represent complex graphs. However, it will be very challenging to interpret those embeddings compared to the original graph structures. While most graph-based methods use structural patterns to detect anomalies, it is difficult to explain which part of graph structure (nodes, edges, or subgraphs) is important for anomaly detection. Thus, it is essential to enhance the interpretability and explainability of graph-based anomaly detection by explaining the graph feature importance, visualizing learned embeddings or weights, or incorporating human validation, for example. 

\subsection{Adversarial Attacks}
An adversarial attack on a graph-based anomaly detection system is a method that purposely generates adversarial graph data to manipulate anomaly detection mechanisms and evade detection. In graph-based anomaly detection methods, an adversarial attack primarily distorts the graph structure, topology, or attributes to achieve their illegitimate objectives. So, it is important to ensure the reliability and robustness of the graph-based anomaly detection methods when it comes to detecting adversarial attacks. By incorporating the adversarial produced data into a training set of graph-based anomaly detection models, one can address adversarial attack difficulties and strengthen detection models. Additionally, one could develop an approach that distinguishes between adversarial anomalies and real anomalies. Furthermore, one can evaluate a graph-based anomaly detection system to determine the impact of adversarial perturbations on graph data.
\section{Conclusion}\label{con}
In this survey, we provide a comprehensive overview of graph-based anomaly techniques. We categorized the anomaly detection techniques and their related papers into four categories: anomalous component, graph type, method applied, and anomaly type, along with several sub-categories within each type. This study consolidates different graph-based anomaly detection papers and briefly discusses the application domains, and provides links to network datasets or repositories used by different graph-based anomaly detection papers. Lastly, this paper underlines some challenges faced by the current graph-based anomaly detection models and potential insights to expand future research directions.

\section*{Dataset Links}
{\customsize
\indent\numberedlink{https://www.cs.cmu.edu/\~./enron}{https://www.cs.cmu.edu/\~./enron}\\
\indent\numberedlink{https://www.yelp.com/dataset}{https://www.yelp.com/dataset}\\
\indent\numberedlink{https://dblp.uni-trier.de/xml/}{https://dblp.uni-trier.de/xml/}\\
\indent\numberedlink{https://www.ll.mit.edu/r-d/datasets/1999-darpa-intrusion-detection-evaluation-dataset}{https://www.ll.mit.edu/r-d/datasets/1999-darpa-intrusion-detection-evaluation-dataset}\\
\indent\numberedlink{https://snap.stanford.edu/data/email-EuAll.html}{https://snap.stanford.edu/data/email-EuAll.html}\\
\indent\numberedlink{https://www.nyc.gov/site/tlc/about/tlc-trip-record-data.page}{https://www.nyc.gov/site/tlc/about/tlc-trip-record-data.page}\\
\indent\numberedlink{https://research.unsw.edu.au/projects/unsw-nb15-dataset}{https://research.unsw.edu.au/projects/unsw-nb15-dataset}\\
\indent\numberedlink{https://www.stratosphereips.org/datasets-ctu13}{https://www.stratosphereips.org/datasets-ctu13}\\
\indent\numberedlink{https://snap.stanford.edu/data/CollegeMsg.html}{https://snap.stanford.edu/data/CollegeMsg.html}\\
\indent\numberedlink{https://networkrepository.com/soc-digg.php}{https://networkrepository.com/soc-digg.php}\\
\indent\numberedlink{https://snap.stanford.edu/data/cit-HepTh.html}{https://snap.stanford.edu/data/cit-HepTh.html}\\
\indent\numberedlink{https://networkrepository.com/ca-IMDB.php}{https://networkrepository.com/ca-IMDB.php}\\
\indent\numberedlink{https://kdd.ics.uci.edu/databases/kddcup99/kddcup99.html}{https://kdd.ics.uci.edu/databases/kddcup99/kddcup99.html}\\
\indent\numberedlink{https://odds.cs.stonybrook.edu/twittersecurity-dataset/}{https://odds.cs.stonybrook.edu/twittersecurity-dataset/}\\
\indent\numberedlink{https://anlab-kaist.github.io/traces/}{https://anlab-kaist.github.io/traces/}\\
\indent\numberedlink{https://datahub.hku.hk/articles/dataset/Weiboscope\_Open\_Data/16674565}{https://datahub.hku.hk/articles/dataset/Weiboscope\_Open\_Data/16674565}\\
\indent\numberedlink{https://odds.cs.stonybrook.edu/twitterworldcup2014-dataset/}{https://odds.cs.stonybrook.edu/twitterworldcup2014-dataset/}\\
\indent\numberedlink{https://snap.stanford.edu/data/soc-Epinions1.html}{https://snap.stanford.edu/data/soc-Epinions1.html}\\
\indent\numberedlink{https://linqs.org/datasets/\#cora}{https://linqs.org/datasets/\#cora}\\
\indent\numberedlink{https://linqs.org/datasets/\#pubmed-diabetes}{https://linqs.org/datasets/\#pubmed-diabetes}\\
\indent\numberedlink{https://linqs.org/datasets/\#citeseer-doc-classification}{https://linqs.org/datasets/\#citeseer-doc-classification}\\
\indent\numberedlink{https://linqs.org/datasets/\#webkb}{https://linqs.org/datasets/\#webkb}\\
\indent\numberedlink{https://github.com/snap-stanford/ogb}{https://github.com/snap-stanford/ogb}\\
\indent\numberedlink{http://konect.cc/networks/arnetminer/}{http://konect.cc/networks/arnetminer/}\\
\indent\numberedlink{https://www.unb.ca/cic/datasets/index.html}{https://www.unb.ca/cic/datasets/index.html}\\
\indent\numberedlink{https://snap.stanford.edu/data/Oregon-2.html}{https://snap.stanford.edu/data/Oregon-2.html}\\
\indent\numberedlink{http://download.srv.cs.cmu.edu/\~mmcgloho/fec/data/fec\_data.html}{http://download.srv.cs.cmu.edu/\~mmcgloho/fec/data/fec\_data.html}\\
\indent\numberedlink{https://snap.stanford.edu/data/soc-RedditHyperlinks.html}{https://snap.stanford.edu/data/soc-RedditHyperlinks.html}\\
\indent\numberedlink{https://snap.stanford.edu/data/sx-stackoverflow.html}{https://snap.stanford.edu/data/sx-stackoverflow.html}\\
\indent\numberedlink{https://snap.stanford.edu/data/web-Amazon.html}{https://snap.stanford.edu/data/web-Amazon.html}\\
\indent\numberedlink{https://snap.stanford.edu/data/wiki-Vote.html}{https://snap.stanford.edu/data/wiki-Vote.html}\\
\indent\numberedlink{https://www.cs.cmu.edu/~kijungs/codes/kcore/}{https://www.cs.cmu.edu/~kijungs/codes/kcore/}\\
\indent\numberedlink{https://www.cs.umd.edu/hcil/VASTchallenge08/}{https://www.cs.umd.edu/hcil/VASTchallenge08/}\\
\indent\numberedlink{https://www.barracuda.com/}{https://www.barracuda.com/}\\
\indent\numberedlink{https://sparse.tamu.edu/}{https://sparse.tamu.edu/}\\
\indent\numberedlink{https://www.fruitfly.org/}{https://www.fruitfly.org/}\\
\indent\numberedlink{https://www.lbl.gov/}{https://www.lbl.gov/}\\
\indent\numberedlink{https://www.icir.org/enterprise-tracing/}{https://www.icir.org/enterprise-tracing/}\\
\indent\numberedlink{https://internet2.edu/community/research-engagement/research-funding-support/}{https://internet2.edu/community/research-engagement/research-funding-support/}\\
\indent\numberedlink{https://dublin.sciencegallery.com/infectious}{https://dublin.sciencegallery.com/infectious}\\
\indent\numberedlink{https://pems.dot.ca.gov/}{https://pems.dot.ca.gov/}\\
\indent\numberedlink{http://realitycommons.media.mit.edu/realitymining.html}{http://realitycommons.media.mit.edu/realitymining.html}\\
\indent\numberedlink{https://www.cs.cmu.edu/~kijungs/codes/mzoom/}{https://www.cs.cmu.edu/~kijungs/codes/mzoom/}\\
\indent\numberedlink{https://www.cs.cmu.edu/~kijungs/codes/dcube/}{https://www.cs.cmu.edu/~kijungs/codes/dcube/}\\
\indent\numberedlink{http://www3.cs.stonybrook.edu/~emanzoor/streamspot/}{http://www3.cs.stonybrook.edu/~emanzoor/streamspot/}\\
\indent\numberedlink{https://snap.stanford.edu/data/index.html}{https://snap.stanford.edu/data/index.html}\\
\indent\numberedlink{http:reality.media.mit.edu/}{http:reality.media.mit.edu/}\\
\indent\numberedlink{http://konect.cc/networks/}{http://konect.cc/networks/}\\
\indent\numberedlink{https://infolab.tamu.edu/data/}{https://infolab.tamu.edu/data/}\\
\indent\numberedlink{https://snap.stanford.edu/data/web-Movies.html}{https://snap.stanford.edu/data/web-Movies.html}\\
\indent\numberedlink{https://snap.stanford.edu/biodata/datasets/10000/10000-PP-Pathways.html}{https://snap.stanford.edu/biodata/datasets/10000/10000-PP-Pathways.html}\\
\indent\numberedlink{https://datasets.syr.edu/datasets/BlogCatalog.html}{https://datasets.syr.edu/datasets/BlogCatalog.html}\\
\indent\numberedlink{https://datasets.syr.edu/datasets/Flickr.html}{https://datasets.syr.edu/datasets/Flickr.html}\\
\indent\numberedlink{https://github.com/THUDM/GraphCAD}{https://github.com/THUDM/GraphCAD}\\
\indent\numberedlink{https://github.com/PanpanZheng/OCAN/tree/master/data}{https://github.com/PanpanZheng/OCAN/tree/master/data}\\
\indent\numberedlink{https://github.com/d-ailin/GDN/tree/main}{https://github.com/d-ailin/GDN/tree/main}\\
\indent\numberedlink{http://networkrepository.com/email-dnc}{http://networkrepository.com/email-dnc}\\
\indent\numberedlink{http://snap.stanford.edu/data/soc-sign-bitcoin-alpha}{http://snap.stanford.edu/data/soc-sign-bitcoin-alpha}\\
\indent\numberedlink{http://snap.stanford.edu/data/soc-sign-bitcoin-otc}{http://snap.stanford.edu/data/soc-sign-bitcoin-otc}\\
\indent\numberedlink{http://networkrepository.com/tech-as-topology}{http://networkrepository.com/tech-as-topology}\\
\indent\numberedlink{http://www.ipd.kit.edu/˜muellere/consub/}{http://www.ipd.kit.edu/˜muellere/consub/}\\
\indent\numberedlink{https://snap.stanford.edu/jodie/\#datasets}{https://snap.stanford.edu/jodie/\#datasets}\\
\indent\numberedlink{http://realitycommons.media.mit.edu/socialevolution.html}{http://realitycommons.media.mit.edu/socialevolution.html}\\
\indent\numberedlink{https://www.gharchive.org/}{https://www.gharchive.org/}\\
\indent\numberedlink{http://networkrepository.com/fb-wosn-friends.php}{http://networkrepository.com/fb-wosn-friends.php}\\
\indent\numberedlink{http://snap.stanford.edu/data/sx-mathoverflow.html}{http://snap.stanford.edu/data/sx-mathoverflow.html}\\
\indent\numberedlink{https://chrsmrrs.github.io/datasets/docs/datasets/}{https://chrsmrrs.github.io/datasets/docs/datasets/}\\
\indent\numberedlink{ https://www.scholat.com/}{ https://www.scholat.com/}\\
\indent\numberedlink{https://www.aminer.org/data/}{https://www.aminer.org/data/}\\
\indent\numberedlink{https://rose1.ntu.edu.sg/dataset/actionRecognition/}{https://rose1.ntu.edu.sg/dataset/actionRecognition/}\\
\indent\numberedlink{https://archive.org/details/asin\_listing/}{https://archive.org/details/asin\_listing/}\\
\indent\numberedlink{https://networkrepository.com/index.php}{https://networkrepository.com/index.php}\\
\indent\numberedlink{https://snap.stanford.edu/index.html}{https://snap.stanford.edu/index.html}\\
\indent\numberedlink{https://www.ll.mit.edu/r-d/datasets}{https://www.ll.mit.edu/r-d/datasets}\\
\indent\numberedlink{https://infolab.tamu.edu/data/}{https://infolab.tamu.edu/data/}\\
\indent\numberedlink{http://konect.cc/}{http://konect.cc/}\\
\indent\numberedlink{https://www.stratosphereips.org/datasets-overview}{https://www.stratosphereips.org/datasets-overview}\\
\indent\numberedlink{https://anlab-kaist.github.io/traces/}{https://anlab-kaist.github.io/traces/}\\
\indent\numberedlink{https://datasets.syr.edu/pages/datasets.html}{https://datasets.syr.edu/pages/datasets.html}\\
\indent\numberedlink{https://www.caida.org/}{https://www.caida.org/}\\

}

\bibliographystyle{IEEEtran}
\bibliography{main.bib}
\end{document}